\documentclass{article} %
\usepackage{iclr2027_conference,times}

\usepackage{amsmath,amsfonts,bm}

\def\eqref#1{equation~\ref{#1}}

\def\1{\bm{1}}

\DeclareMathAlphabet{\mathsfit}{\encodingdefault}{\sfdefault}{m}{sl}
\SetMathAlphabet{\mathsfit}{bold}{\encodingdefault}{\sfdefault}{bx}{n}

\usepackage{hyperref}
\usepackage{url}
\usepackage{hyperref}
\usepackage{url}
\usepackage{amsmath}
\usepackage{amsthm}
\usepackage{amssymb}
\usepackage{graphicx}
\usepackage{listings}
\usepackage{float}
\usepackage{booktabs}
\usepackage{xcolor}
\usepackage{enumitem}
\usepackage{subcaption}
\usepackage{longtable}
\usepackage{fontawesome}

\newcommand{\tbAccWIFUned}{0.28}

\newcommand{\tbAccWIFCtrl}{0.29}

\newcommand{\tbAccWCLoo}{0.99}

\newcommand{\tbAccDTFSev}{0.08}

\newcommand{\tbAccDCLoo}{0.09}

\newcommand{\tbAccWTFPromptFBel}{0.99}

\newcommand{\tbAccDTFLooFBel}{0.33}

\newcommand{\toyRefF}{0.26}
\newcommand{\toyRefL}{0.125}
\newcommand{\toyRefFL}{0.032}

\title{LLMs program using Reusable \\and 
Composable Abstract Representations}

\author{Casper L. Christensen\thanks{Correspondence: \texttt{casper.lutzhoft@gmail.com}} \\
Simplex \\
Astera Institute \\
\And
Kyle J. Ray \\
Simplex \\
Astera Institute
\And
Dmitry Manning-Coe \\
MATS \\
University of Illinois at Urbana-Champaign
\AND
Adam Shai \\
Simplex \\
Astera Institute
\And
Paul Riechers \\
Simplex \\
Astera Institute
}

\iclrfinalcopy %
\begin{document}

\maketitle
\begin{abstract}
When writing code, the same computation can be expressed in different programming languages and through different implementation strategies. We show that the activations of LLMs capture this shared structure in reusable and composable abstract representations when they generate code. These representations emerge in the middle layers, are reusable across programming languages, and afford the model an impressive amount of generalization ability when generating code under various interventions. Using activation steering, we show that function, programming language, and implementation strategy can be independently manipulated. The independent interventions on language and function can be flexibly composed to control code generation, even in cases when the prompt contains no information about language and function. We additionally find that finetuning on examples of arbitrarily defined functions in one programming language transfers to other languages, consistent with the models reusing abstract features of code generation that are shared across different examples. Finally, to study how such abstract representations come to be, we implement a minimal model of shared, reusable, and composable structure in a synthetic setting, and show that transformers trained on such data learn representations that reuse these parts flexibly across contexts.
\end{abstract}

\section{Introduction}
Programming repeatedly combines familiar computations with different languages and implementation choices. A graph search, for example, can recur across programs with different syntax, control flow, and data structures. \textbf{A competent programmer can recognize what these implementations share, and reason about the abstract computation, separately from the details of its realization.} This allows knowledge acquired in one setting to support programming in another; understanding a graph-search algorithm in Python can help a programmer implement it in Java, even when the implementation requires different language constructs and data structures. Rather than learning every realization independently, programmers can reuse their understanding of the computation while adapting its implementation to the setting. In this work, \textbf{we show that coding models also learn such abstractions from the shared computational structure of program data}. These representations are built from reusable contributions that decompose into 
(i) what a program computes, 
(ii) the language in which it is expressed, and 
(iii) its
implementation strategy. 

Our work builds on the Factored World Hypothesis (FWH), which proposes that \emph{predictive models can represent independent parts of a data-generating process in separable latent state spaces} \citep{factoredworld2026}. Here, we extend the theoretical framework to training data from multiple sources \citep{ray2026nonergodic} that share computational parts. \textbf{When training data from different sources contain shared mechanisms, we anticipate and find the emergence of reusable factored representations}. That is, representations that track information about the shared mechanism abstractly and separately from information about the source-specific components. Representations that leverage such abstractions are more dimensionally efficient at no predictive cost, because the joint representation can be fully recovered from the two marginals: one over the shared mechanism and one over the remaining mixture.

In the following, we ask whether a model's activation state $Z$ (i.e., the residual stream in a transformer \citep{vaswani2017attention}, or the hidden state in a recurrent neural network) learns to exploit this kind of shared structure. We test reuse directly in open-source coding models’ internal representations. To begin, we find that when a programming instruction is put into context, \textbf{the most performant models retain an internal state that supports code generation, even after direct access to the specification is removed}. This compiler-like representation gives us an internal target of study.

Through this compiled state, we ask whether the model represents a function it's been asked to implement separately from the programming language in which it's been asked to write. Our contributions are as follows:

\begin{enumerate}[leftmargin=*,itemsep=0pt,topsep=2pt]
    \item \textbf{Coding LLMs create a compiled program state representation.} Across 26 instruction-tuned models, the strongest coders keep enough task information at a shared suffix to write correct code after the specification is masked (Section~\ref{sec:compiler}).
    \item \textbf{LLMs represent function, language, and strategy as reusable, composable components.} Interventions estimated only from other languages (a) transport, (b) install, and (c) compose (i) function, (ii) language, and (iii) strategy in three LLMs and a modern RNN (Section~\ref{sec:llm-factors}).
    \item \textbf{Training on data with shared mechanisms produces reusable representations.} In a controlled synthetic setting, next-token prediction on data whose sources share a mechanism suffices to produce reusable, composable structure (Section~\ref{sec:toy}).
    \item \textbf{Reusable abstractions let models generalize across languages.} State from a single in-context example transfers across languages, and Python-only fine-tuning on invented algorithms transfers to held-out languages (Section~\ref{sec:icl}).
\end{enumerate}
Code to reproduce all experiments is available at 
\faicon{github}~~\url{https://github.com/Astera-org/code-abstractions}.

\begin{figure}
    \centering
    \includegraphics[width=0.99\linewidth]{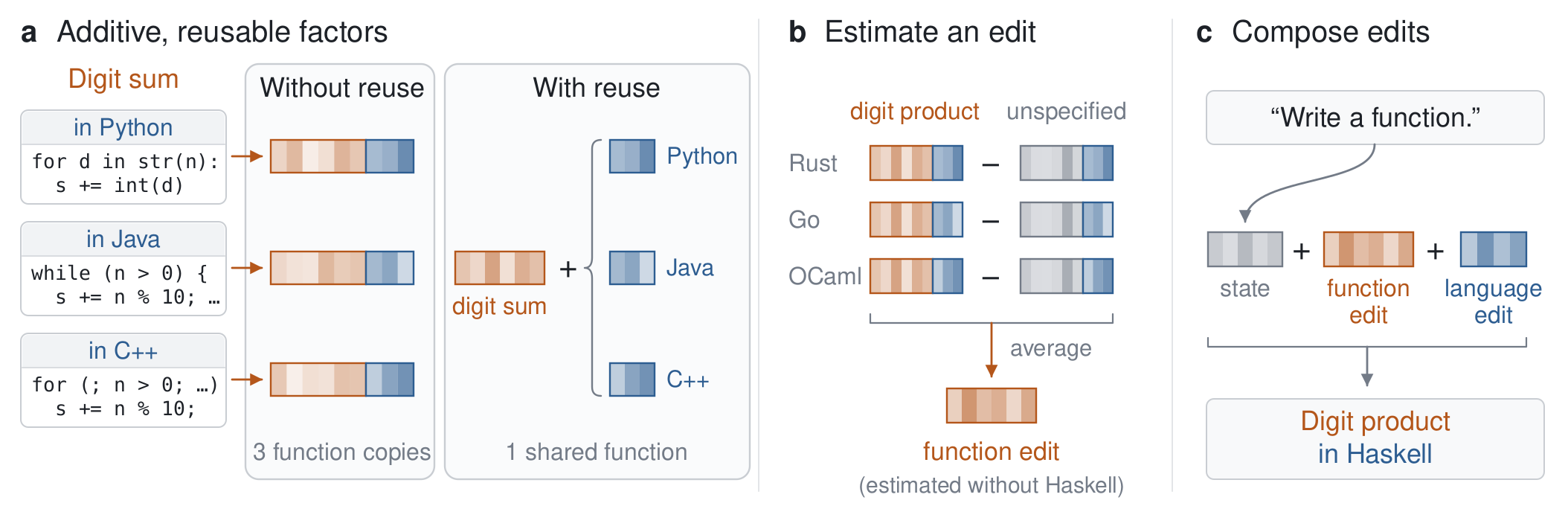}
    \caption{\textbf{Abstract representations enable control and generalization in programming.} \textbf{(a)}: The factored world hypothesis motivates additive activation states; when multiple factored sources reuse the same mechanism, a reusable representation avoids redundancy. \textbf{(b,c)}: Testing composability and reusability of coding representations. In (b), we find a steering vector for a fixed function without ever asking the model to output Haskell code (separately, we also find a steering vector for Haskell, without ever asking the model to implement the target function.) In (c) we use both steers at once to implement the target function in Haskell.}
    \label{fig:concept}
\end{figure}

\section{Related Work}

\paragraph{Shared representations and Cross-Language Generalization} Prior work has found semantic structure in code-model representations, including emergent program semantics \citep{jin2024semantics}, language-agnostic code embeddings \citep{utpala2024language}, and shared conceptual representations across natural languages \citep{dumas2025separating}. Other work directly encourages abstraction, for example, by training models against compiler intermediate representations \citep{paul2024ircoder}. Similarly, cross-programming-language transfer is well established behaviorally. Language models often transfer knowledge between languages, with reliability depending on the source language, target language, and task \citep{baltaji2025crosslingual}. Parallel-SFT shows that explicitly aligning equivalent implementations across languages can improve zero-shot transfer during subsequent code training \citep{wu2026parallel}, while other work studies what syntax and semantics transfer when models encounter unfamiliar programming languages \citep{kumar2026syntax}. We directly extend this work with an in-depth representational study, asking if the representations naturally learned during pretraining are factored abstractions and show that behavior like cross-language transfer may be a natural consequence of a factored, reusable representation.

\paragraph{Task and function representations.} Task vectors can summarize demonstrations \citep{hendel2023taskvectors}, while function vectors localized to attention-head activity can causally induce learned input-output mappings and exhibit limited compositionality \citep{todd2024functionvectors}. In-context vectors provide a related latent-space intervention framework \citep{liu2024incontext}. Recent work has further asked whether these representations are invariant to how a task is presented. \citet{davidson2025prompting} find that instructions and demonstrations rely largely on distinct, partially overlapping function-vector mechanisms, whereas \citet{luo2025crossmodal} find task representations that transfer across modality and presentation format. Here, we use representations obtained from language models that correspond directly to programming languages, functions, and implementation details, and show that these can in general be transferred and composed to induce new computation.

\paragraph{Activation and cache interventions.}
Latent steering vectors, Activation Addition, and Contrastive Activation Addition manipulate generation using activation-space directions derived from desired behaviors or contrastive prompts \citep{subramani2022steering,turner2023actadd,rimsky2024caa}; instruction vectors can similarly be composed to control multiple behavioral constraints \citep{stolfo2025instruction}. Code-specific work has used activation steering for properties such as types, language, and library choice \citep{lucchetti-guha-2025-understanding,rahman2026steering}. Prefix-tuning and KV-cache steering demonstrate that persistent prompt or cache state can also exert substantial control over subsequent generation \citep{li2021prefix,belitsky2025kvcache}. In our work we use steering as our intervention of choice to causally test the predictions a factored, reusable representation makes.

\paragraph{Factoring.} An intuition has been building that neural networks often seem to factor low-level syntax from abstract semantics.  This is seen in image generation, where the style of a painting can be altered independently of its content \citep{gatys2015neural}.  And it's seen in language models, where syntax and semantics factor both at the level of particular words \citep{feucht2025vector} and at the level of full sentences \citep{acevedo2026differential}.

Our work lends further credence to this intuition, while providing a very sharp set of examples via compositional programming representations, and supplying a toy model that helps explain where this factored abstract structure can come from.

\section{Factored Program State and Causal Tests}
\label{sec:program-data-model}

Programming tasks provide a natural example of a structured, non-ergodic data source \citep{ray2026nonergodic}. Computation recurs across many otherwise different contexts. For example, a graph search can be implemented in Python or C++, recursively or iteratively, and using depth-first or breadth-first traversal. We model these recurring structures with task variables $\theta=(f,\lambda,s)$, where $f$ denotes the computation or input--output mapping, $\lambda$ the programming language, and $s$ the implementation strategy. We can think of a program $u$ as being drawn from a task-conditioned source.
\begin{equation}
    u\sim p(u\mid\theta).
    \label{eq:program-generative-model}
\end{equation}
Conditioning on a task $\theta$ means supplying a prompt that specifies $f$, $\lambda$, and $s$. Language can take different values, such as ``Python'' or ``Java'', while keeping one or both of function and implementation fixed, and likewise for the other variables.

The Factored World Hypothesis \citep{factoredworld2026} motivates a decomposition of some \textbf{persistent \emph{activation state} $Z(\theta)$ into separable, additive contributions associated with function, language, and strategy}. We denote these contributions by $Z_F(\theta),Z_L(\theta),Z_S(\theta)$.

Our \emph{reuse hypothesis} further proposes that \textbf{each contribution depends primarily on its own variable, rather than the entire task}. For language, we express this as $Z_L(\theta)\approx Z_L(\lambda)$:
the same language representation should be shared across functions and strategies. Applying the reuse assumption to function and strategy gives $Z$ the form
\begin{equation}
    Z(\theta) \approx Z_0+Z_F(f)+Z_L(\lambda)+Z_S(s).
    \label{eq:code-factorization}
\end{equation}
Here $Z_0$ denotes whatever part of the activation state is shared by all tasks. Some decomposition of this form is predicted to hold for each task, though it may be subject to compatibility constraints that restrict which combinations can occur.

We test the operational prediction that \emph{individual factor changes} transfer across dissimilar contexts: a state change intervention
estimated in one context should produce the corresponding change in another. For example, an edit estimated from analogous tasks in non-Python
languages should be able to change the state for summing digits in Python into one that produces their
product in Python. This prediction follows directly from Eq.~(\ref{eq:code-factorization}), because
\begin{equation}
      Z(f^\star,\lambda,s)-Z(f,\lambda,s) \approx Z_F(f^*)-Z_F(f).
    \label{eq:transfer-prediction}
\end{equation}
The final displacement vector depends on $f$, not the source language or implementation strategy. Here, the activation state need not equal a Bayesian belief or inherit tensor-product coordinates. The hypothesis we test
follows from, but does not require, orthogonal subspaces. Nor does it depend on the pressure to minimize representational dimension. In fact, it does not even require that the factoring holds at every layer of a model. 

\textbf{Testing this prediction does require identifying a reliable activation state $Z(\theta)$} through which the model can represent, and we can modify, the task specification. In the next section, we empirically find a candidate site to hold such abstractions in LLMs. Then, we use it to investigate causal signatures of representational reuse using natural language programming prompts. We then show the same behavior in a toy process with an exactly known data distribution in which theory predicts such a representation.

\subsection{Modern Models Exhibit Compiler-Like Prompt State}
\label{sec:compiler}

We look for an intervention site by asking whether a model retains useful task information after direct access to the program specification tokens is removed. Each prompt contains a task-bearing prefix followed by a shared suffix. During prefill, \textbf{we process the full prompt and cache the key--value states. We then
mask attention to the specification positions}, denoted $\mathcal D$, and generate a program. This tests whether information propagated beyond the specification token positions can support the task; prompt details are in Appendix~\ref{app:prompting}.

We screen 26 instruct-tuned LLMs on 310 coding tasks spanning 10 languages, then repeat the task masking out attention to the prefix. Success requires passing all unit tests while only attending to the shared suffix.
(Appendices~\ref{app:llms} and~\ref{app:programs}).

\begin{figure}[H]
    \centering
    \includegraphics[width=0.99\linewidth]{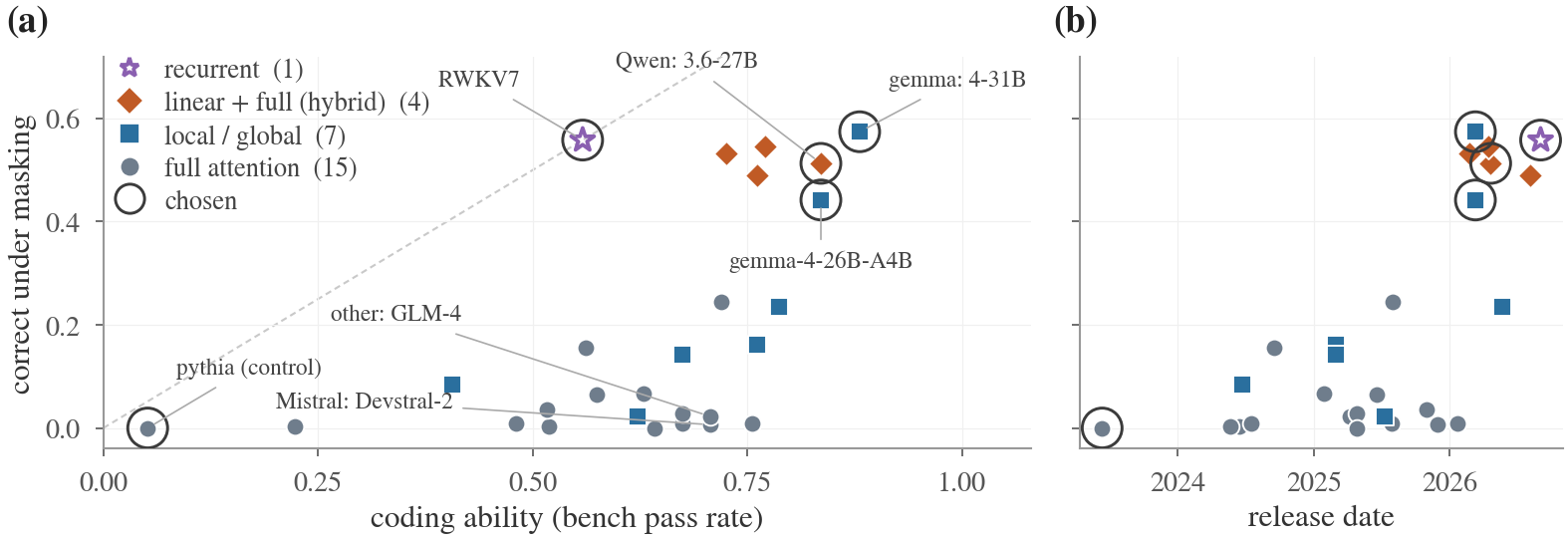}
    \caption{\textbf{Modern models compile coding specifications.} \textbf{(a)}: The most capable coding models are more resistant to not being allowed to attend to the description after prefill. These models also mix local or linear attention with global, or are fully recurrent. \textbf{(b)}: This also largely correlates with the recency of the model.}
    \label{fig:compiler-like}
\end{figure}

The most capable coding models retain substantial performance under this intervention, indicating that \textbf{the task information provided in the prefix is ``compiled'' into a representation that exists in the activations at the shared suffix context positions}.
(Fig.~\ref{fig:compiler-like}). This provides a controllable local site that we use as our $Z(\theta)$ for testing whether the function and language contributions can be reused separately. We now proceed to test if this $Z(\theta)$ behaves as predicted in Eq.~(\ref{eq:code-factorization}).

\subsection{Intervention Design}
\label{sec:prompt-setup}
\label{sec:shared-interventions}

To compare states across tasks, we build all prompts using the same format. For task $\theta$, the full prompt is
\begin{equation}
    x= \tau_\theta \, \Vert \, \sigma ~,
    \label{eq:prompt-structure}
\end{equation}
where $\tau_\theta$ is a task-bearing prefix, $\sigma$ a shared suffix, and $\Vert$ token concatenation. $\tau_\theta$ is constructed to ensure that all prompts reach the suffix at the same token positions, so differences in the activations induced by the suffix context positions reflect only information propagated from the prefix. We call prompts used to estimate an edit \emph{donors} and the prompt whose state receives the edit the \emph{recipient}. Natural language prompt construction details and examples can be found in Appendix~\ref{app:prompting}.

After prefill, the editable key--value cache block is $Z(\tau_\theta)=\{K_{\ell,i},V_{\ell,i}\}_{\ell\in\mathcal L,\,i\in\mathcal S}$ where $K_{\ell,i}$ and $V_{\ell,i}$ are the key and value tensors at layer $\ell$ and position $i$;
$\mathcal L$ selects layers and $\mathcal S$ suffix positions. In each intervention, we apply $\widetilde Z=Z+\alpha\Delta,$ with displacement $\Delta$ and dose $\alpha$, matching each layer, position, and key or value entry. Description-access removal is a separate operation, specified for each experiment---but effectively amounts to masking the attention mechanism so that it may only attend to the shared, intervened suffix representations formed while the description was visible.

Under the reuse hypothesis, contrasting two functions in the same language cancels the language contribution. Write $\overline Z(\theta)$ for the mean
suffix state over prompts that realize $\theta$.
An edit from source function $f$ to target $f^\star$ is estimated by
\begin{equation}
    \Delta_F[f^\star\leftarrow f]
    =\underset{\lambda'\in\Lambda}{\operatorname{mean}}
      \left[\overline Z(f^\star,\lambda')-\overline Z(f,\lambda')\right].
    \label{eq:loo-factor}
\end{equation}
Here $\Lambda$ is the set of languages whose prompts serve as donors. When $\Lambda$ leaves out the recipient's language, the edit is built entirely from other languages, so success shows that the factor transfers across contexts; we call such a test a ``leave-one-context-out'' intervention. When $\Lambda$ contains only the recipient's language, we call it a ``same-context-only'' intervention. A language edit, $\Delta_L$, analogously averages contrasts across various functions.

\paragraph{Three tests of reuse.}
 We will test increasingly demanding uses of the same estimator. Below, ``$?$'' indicates a task prompt where one or more of the factors are left unspecified.
\begin{itemize}[itemsep=0pt]
    \item \textbf{Transport:} change $(f,\lambda)$ to $(f^\star,\lambda)$ using
    $\Delta_F[f^\star\leftarrow f]$, preserving the recipient language.
    \item \textbf{Installation:} start from $(?,\lambda)$ and supply the target function using
    $\Delta_F[f^\star\leftarrow ?]$.
    \item \textbf{Composition:} start from $(?,?)$ and add separately estimated function and language
    installation edits to obtain $(f^\star,\lambda^\star)$.
\end{itemize}
Below, we report the fraction of recipients that pass tests designed for the target prompt. Appendix~\ref{app:notation} gives the full estimator and intervention equations. Each experiment below defines the specific ways in which this overarching approach applies.

\section{Testing Transport, Install, and Compose}
\subsection{Language Models Reuse Function, Language, and Strategy}
\label{sec:llm-factors}
\begin{figure}[H]
    \centering
    \includegraphics[width=0.99\linewidth]{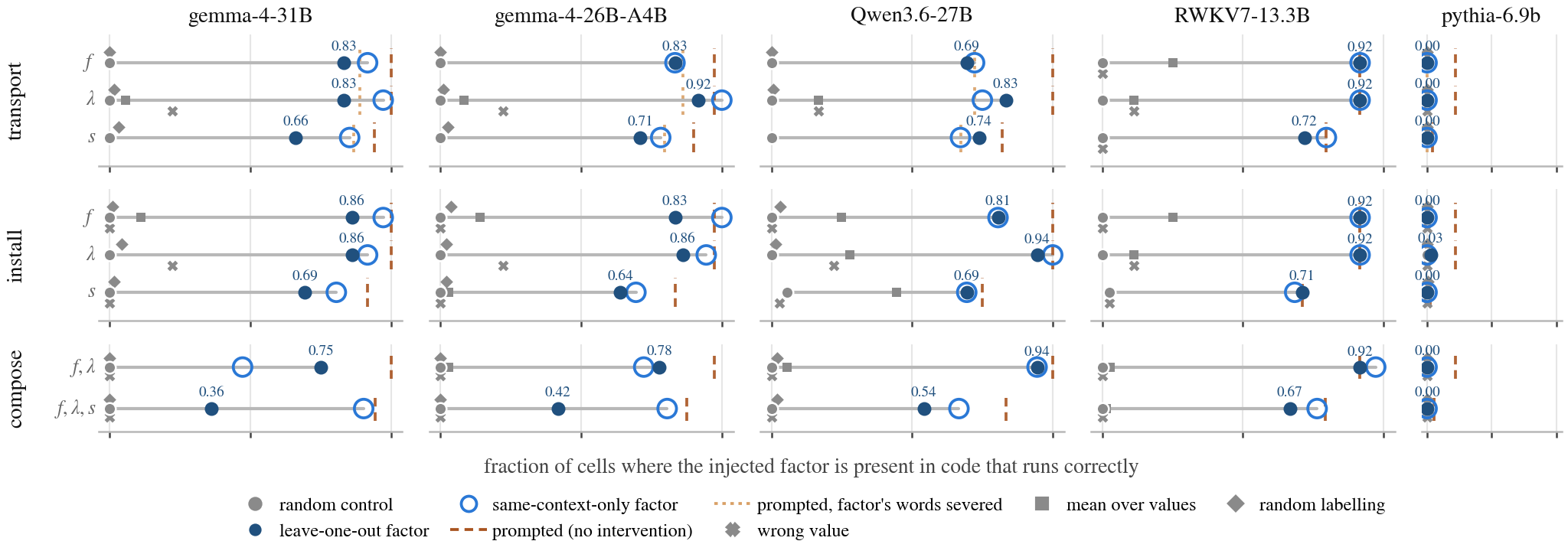}
    \caption{\textbf{Function, language, and strategy support causal reuse.}
     For each condition, we report the
    fraction of evaluation cells in which the generated program both runs
    correctly and exhibits the injected factor. Function and
    language factors estimated outside the recipient context generally recover
    most of the effect of same-context-only interventions, and strategy factors likewise causally control recursive versus iterative realization. Function--language composition remains strong, while simultaneous composition of function,
    language, and strategy is substantially more difficult. The Pythia control
    shows no corresponding causal reuse. Interventions with $s$ have 72 trials, all others have 36 (since Python is never a recipient).
    }
    \label{fig:cfs}
\end{figure}

From the retention screen in Section~\ref{sec:compiler}, we select the top three models, \verb|Gemma-4-31B|, \verb|Gemma-4-26B-A4B| \citep{team2026gemma}, and \verb|Qwen3.6-27B| \citep{bai2023qwen}. This also gives us a regular local/global attention transformer, a Mixture-of-Experts (MoE), and a hybrid model with linear and regular attention. We carry \verb|Pythia-6.9b| \citep{biderman2023pythia} through the rest of the experiments as a control, since it has not been extensively trained on code, it lets us validate that our manipulations do not trivially work in any language model. Additionally, we perform all interventions on \verb|RWKV7-13.3B| \citep{peng2025rwkv}, due to its recurrent state giving us an exact place to intervene since there is no question of where its representation exists.

We apply the transport, installation, and composition tests of Section~\ref{sec:shared-interventions} to a subset of 4 tasks the models \textit{are} able to solve without access to the description during generation in 10 different programming languages. For transport and installation, after prefill, \textbf{we mask attention to the description tokens, forcing the model to generate entirely based on its compiled state}. Qwen's recurrent memory also receives the donor-derived edit. Appendix~\ref{app:prompting} and Appendix~\ref{app:programs} contain full intervention details and choice of tasks and languages.

Strategy extends the same estimator to contrasts between implementations, averaged over other
function--language contexts. In our case, we define strategy to be the choice between iteration or recursion. We test function--language composition and the harder combination of all
three factors. We consider an intervention a success when the decoded program passes 6 unit tests. Note that we always leave out \verb|Python| as a recipient since this is every tested model's default language and would inflate our score.

Figure \ref{fig:cfs} shows that the causal predictions of factorization hold across the tested models. Filled blue markers use leave-one-context-out estimates; open blue markers use same-context-only estimates. Gray markers show control interventions, and dashed orange lines show ordinary prompting without intervention. Together, these results show that \textbf{function, language, and strategy are causally reusable components of the compiled program state}. In our design, the donor and recipient share exact wording of each description, which likely contributes to the high success rate. In Appendix~\ref{app:paraphrase} we show additional results where donors and recipients are paraphrases of one another.

It is worth noting that we chose this cast of models because they share an addressable $Z(\theta)$ site at a shared suffix and our interventions are designed accordingly. That the interventions fail for other models does not mean that they also do not make use of similarly composable abstract representations. Rather, it suggests we would have to address them at a different interface if they do. We touch on this question in Appendix~\ref{app:addressability}, but leave a more careful analysis for future work.

\subsection{How could this be happening?}
\label{sec:toy}
What structure in training data makes this kind of shared representation useful? We propose that it is data in which many sources reuse a common mechanism, like how programs in different languages reuse the same computations. In this section, we show  why this type of data favors a representation that stores the shared mechanism once. We then build a small programming-inspired dataset with exactly this structure, train a transformer on its output alone, and apply the transport, installation, and composition tests of Section~\ref{sec:llm-factors}. If next-token prediction alone is enough to produce a reusable factored state, the trained model should show the same causal reuse as the language models in Fig.~\ref{fig:cfs}. Indeed this is what we find.

Consider data from many sources that share a mechanism, such as the same function (say, $x^2$) written in different programming languages (say, Python or C). A network can represent the shared mechanism separately inside each source, or once for all of them (Fig.~\ref{fig:concept}a). Both support the same next-token predictions, but representing it once needs fewer dimensions, and the saving grows linearly with the number of sources (with $K$ sources, representing it once saves $K-1$ copies of the shared mechanism).\footnote{This relies on the FWH result \citep{factoredworld2026} that networks can represent a product of factors as a direct sum of their marginals. See Appendix~\ref{app:shared-mechanism} for more formal detail.}

\begin{figure}[t]
    \centering
    \includegraphics[width=\linewidth]{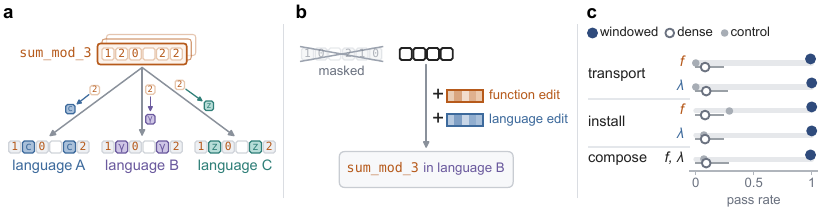}
    \caption{\textbf{A toy training dataset with a shared mechanism.}
    \textbf{(a)} A function (here \texttt{sum\_mod\_3}, one of three) generates a string of symbols (orange). A language turns that string into tokens: each symbol is usually written with a token shared by all languages (gray) and sometimes with a token of that language's own (colored; language A writes 2 as \texttt{c}, language B as $\gamma$). The function is a shared mechanism, analogous to how digit sum is shared by its Python and Rust implementations. The transformer is trained on the tokens alone.
    \textbf{(b)} The composition test of Section~\ref{sec:shared-interventions}: from a prompt that specifies neither factor, we mask the description and add, at the blank suffix, a function edit estimated only from other languages and a language edit estimated only from other functions.
    \textbf{(c)} With an eight-token attention window, the model passes transport, installation and composition in nearly every trial (compose: \tbAccWCLoo). A dense-attention model with the same loss does not (\tbAccDCLoo): even its own prompt, once masked, leaves little at the suffix (\tbAccDTFSev), so accurate prediction alone does not guarantee a reusable state. Leave-one-context-out edits, three seeds (marker: mean; bar: seed range); control is the stronger of the random-direction and wrong-value edits. Full results in Fig.~\ref{fig:toy-cfs}.}
    \label{fig:toy-main}
\end{figure}

Our toy dataset then has the following structure: each sequence comes from one of three \textit{functions} (the analogue of $x^2$), expressed in one of eight \textit{languages} (the analogue of Python or C): the function sets the statistics of the symbols, and the language sets which tokens carry them. The same three functions appear in every language, and the transformer sees only the resulting tokens (Fig.~\ref{fig:toy-main}a; Appendix~\ref{app:toy-process}). Languages also insert runs of blank tokens that carry no task information, which we later use as the shared suffix.

We train transformers with both an eight-position attention window and full dense attention. Both learn the task similarly well ($\approx 8\times10^{-4}$ nats from the optimal loss, averaged over three seeds). The windowed model must relay task information as earlier positions become inaccessible, which creates an incentive for the model to ``compile'' the prompt state in the way we observe in the more competent LLM coders.

Here, $\tau_\theta$ is a token sequence of 36 tokens whose statistics provide evidence about the function and whose tokens indicate the language. A $(?,\lambda)$ prompt omits function information by using a function-ambiguous sequence (one that leaves the optimal observer near a uniform guess), while a $(f,?)$ prompt leaves language ambiguous by using only shared tokens; a $(?,?)$ prompt uses both at once. Here, function specified recipients are selected so that they reveal their task to an optimal Bayesian observer with a certainty of $>95\%$. A series of four ``blank'' tokens forms a suffix shared by all tested contexts, and attention to the 36 token prefix is masked during generation in each trial (Fig.~\ref{fig:toy-main}b).

Because we have an optimal ground-truth distribution of tokens for each prompt, we can determine the success of our interventions by querying the model's next token distribution at just a few positions past the shared suffix. Language, for example, can be read from the probabilities assigned to language-specific tokens immediately after the suffix. Additional training details, prompt construction, filtering, donor counts, and scoring rules are in Appendix~\ref{app:toy-process}. %

Figure~\ref{fig:toy-main}c (full results in Fig.~\ref{fig:toy-cfs}, Appendix~\ref{app:toy-outside}) shows the transport, installation, and composition tests work nearly perfectly for all recipients and donors in the windowed attention model. \textbf{We see that these interventions work equally well whether they use same-context-only or leave-one-out steering---supporting the claim that this toy model has learned composable, reusable representations for language and function}.

The dense attention model achieves similarly low prediction loss but displays much weaker function reuse at the suffix positions. We note that the failure of these interventions merely shows that there is not a reusable representation \emph{at the shared suffix}---not that the dense models do not have a reusable representation. This is consistent with the model's poor performance under prompt masking, even when given the correct target prompt. Additional controls and model architectures are described in Appendix~\ref{app:toy-outside}.

\section{Abstractions support Generalization}
\label{sec:icl}
The preceding experiments show that a specified computation can be realized in different programming languages through a reusable internal state. But the value of an abstraction is not only that it supports different realizations: it should also allow information acquired from one realization to support another. We therefore ask whether reading a single program produces a computational state that can be reused beyond the example’s language. We present one implementation, harvest the persistent state accumulated at a shared suffix, and transplant that state into an otherwise example-free prompt. We pick 2 regular algorithms and 2 ``nonce'' algorithms (arbitrary sequences of steps that do not have a ``common'' name) for each model that exhibits the best performance when attention to the example is masked. Crucially, the target programming language is specified only after the state has been formed, separating acquisition of the computation from the choice of its subsequent realization. We intervene only with programs that successfully survive masking as in Section \ref{sec:shared-interventions}.

\begin{figure}[H]
    \centering

    \begin{subfigure}[t]{0.56\linewidth}
        \centering
        \includegraphics[width=\linewidth]{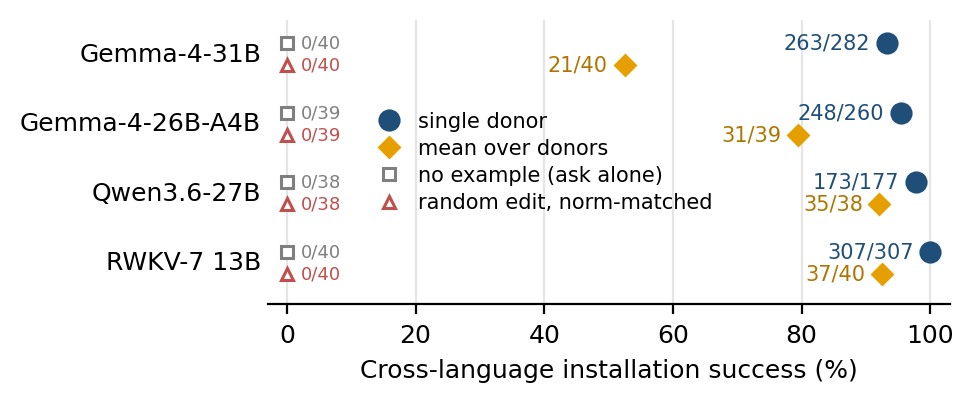}
        \label{fig:icl}
    \end{subfigure}
    \hfill
    \begin{subfigure}[t]{0.43\linewidth}
        \centering
        \includegraphics[width=\linewidth]{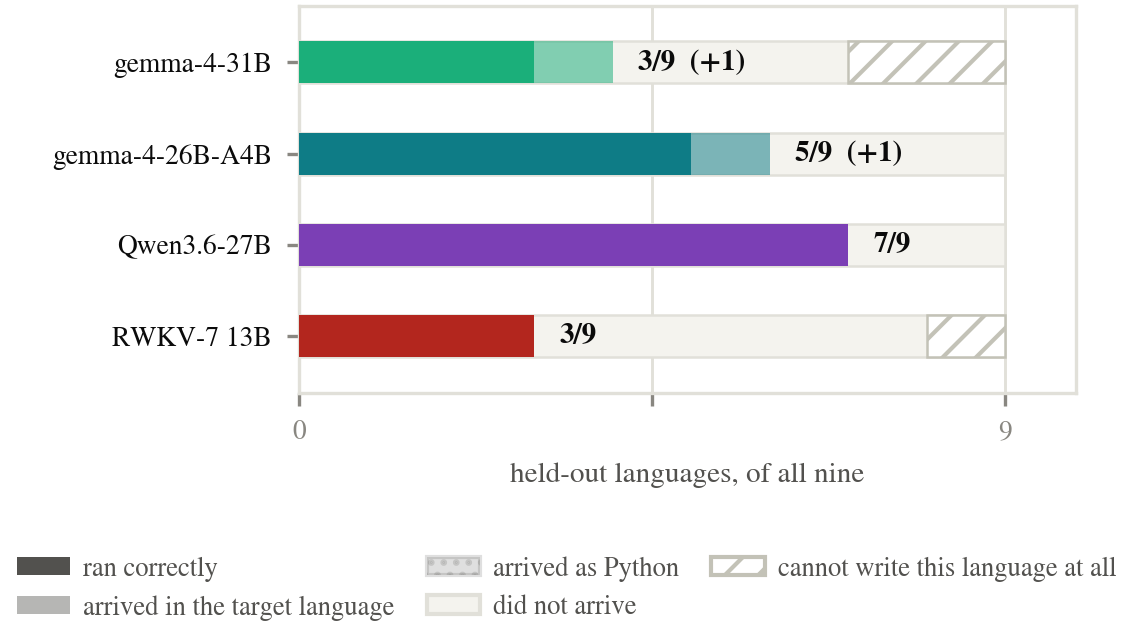}
        \label{fig:sft}
    \end{subfigure}

    \caption{\textbf{A single example provides a transferable representation.} \textbf{(a)}: A state harvested from the suffix prompt with a single example in one language transplanted into an example-free prompt with a different language specified produces a correct program in the new language. \textbf{(b)}: Models trained on a single algorithm in Python learn that algorithm in several held-out languages.}
    \label{fig:icl-sft}
\end{figure}

Figure \ref{fig:icl-sft}a shows that states transplanted from a single prompt with a single language generalize broadly to all other languages. What the model stores at the suffix is thus sufficient for rendering the algorithm in far more languages than it has seen it in. Finally, we try an extreme version of generalization: fine-tune existing models on just a single unseen algorithm in Python to see if they generalize to held-out languages. \citet{wu2026parallel} already showed that training on multiple languages in parallel improves held-out language performance. However, if there is truly a shared representation, a single example should in theory generalize. Figure \ref{fig:icl-sft}b shows the results for an invented algorithm \verb|vantril|. Even in such a data-starved regime, the model manages to generalize to unseen languages. We note that this generalization can likely be considerably improved with more fine-tuning, but consider this beyond the scope of this paper. We expand on these results and describe the full algorithms and experiments in Appendix~\ref{app:icl} and \ref{app:sft}.

\subsection{How abstract are these program representations?}
\label{sec:strategy-abstraction}

We now turn to an examination of what is actually contained in our $Z(\theta)$ state. Previous experiments could succeed if the model simply stored a summary of the prompt or, in the in-context-learning case, the demonstration it saw in its KV-cache. To address this, we look at interventions where the donor is not averaged but comes from a single language.

\begin{figure}[H]
    \centering
    \includegraphics[width=0.99\linewidth]{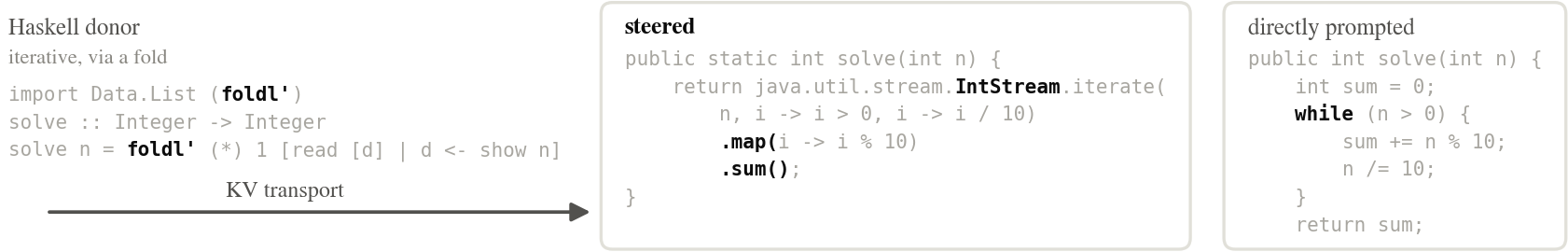}
    \caption{
    \textbf{Cross-language strategy transport can also impart language specific
    information.} Prompting Gemma to write Java, and then steering it with an iterative strategy donor from Haskell induces Java-native syntax: a program using a reduction/stream implementation rather than reproducing Haskell syntax. However, the resulting Java implementation (center) is different from the program that comes from a prompt that directly asks for iteration in Java (right). The steered program is a ``Haskel-like'' approach to writing iteration in Java.
    }
    \label{fig:abstraction}
\end{figure}

Strategy steering provides a nuanced picture of what is contained in $Z_S(\theta$), for example. In most cases, a transported strategy state behaves like a simple label swap in the prompt: when it changes the recipient's program, the result is 
often identical to 
the recipient's own directly prompted program for the target task. However, when the recipient and donor languages are quite different from each other, this does not necessarily hold. Figure~\ref{fig:abstraction} shows a case where steering Java from one strategy to another using a Haskell-derived steer results in a ``Haskell flavored" Java program. So, this steering carries more than a strategy label swap: it is also tied to some donor specific information tied to language. This reinforces our understanding of $Z_S(\theta) \approx Z_S(s)$ as only the dominant behavior. The appropriate strategy for implementing a given function is dependent on the tools available in the language, so a finer grained picture takes this into account, allowing for a correction term $Z_S(\theta) \approx Z_S(s) + \delta Z_S(s,\lambda)$. This is also consistent with strategy interventions working better from the same-context estimate than from the average: the averaging removes the donor-language-specific correction term. We report the full donor $\times$ recipient grid and more examples in Appendix~\ref{app:strategy}.

We then repeat the experiment from Figure \ref{fig:icl-sft} on same-language donor-recipient pairs, and ask how often the generation is identical to the example the model was shown, when attention to it is masked out.

\begin{table}[H]
\caption{The form of the generated program differs when the model is asked to generate without being able to re-read the example and when it is patched.}
\label{tab:samelang}
\begin{center}
\small\setlength{\tabcolsep}{5pt}
\begin{tabular}{l cc cc cc}
\toprule
& \multicolumn{2}{c}{\textsc{visible}} & \multicolumn{2}{c}{\textsc{masked}} & \multicolumn{2}{c}{\textsc{installed}} \\
\cmidrule(lr){2-3}\cmidrule(lr){4-5}\cmidrule(lr){6-7}
\multicolumn{1}{c}{\bf Model} & Identical & Correct & Identical & Correct & Identical & Correct \\
\midrule
Gemma-4-31B & 39/40 & 40/40 & 0/40 & 31/40 & 0/40 & 32/40 \\
Gemma-4-26B-A4B & 35/40 & 40/40 & 1/40 & 30/40 & 1/40 & 30/40 \\
Qwen3.6-27B & 40/40 & 40/40 & 0/40 & 18/40 & 0/40 & 18/40 \\
RWKV-7 13B & 35/40 & 40/40 & --- & --- & 35/40 & 40/40 \\
\bottomrule
\end{tabular}
\end{center}
\end{table}

The results in Table \ref{tab:samelang} are clear. When the model has access to the example, it will repeat it verbatim. However, when forced to generate without being able to re-read the example \textit{or} when being steered, the form of the program often differs. This along with Figure \ref{fig:abstraction} gives a generic picture of what the model stores at its suffix: a program representation that is neither a prompt summary nor a copy of some implementation it has seen. For the algorithms in this experiment, the example is up to 8$\times$ longer than the suffix, so the model also \textit{must} compress the example.

\section{Conclusion}

Just like a competent human programmer, \textbf{LLMs can recognize what distinct code implementations share, and separate the abstract computation from the details of its realization}. They do not relearn coding concepts in various contexts as independent things, but are able to adapt abstractions of those concepts to fit disparate situations. This capability is supported by a compositional state that allows separate and flexible control of 
(i) what a program computes, 
(ii) the language in which it is expressed, and 
(iii) its implementation strategy. 

We show that function, language, and strategy support separable causal interventions through $Z(\theta)$, a ``compiled'' state that is induced by natural language coding prompts and then strongly attended to during generation in LLMs. In-context experiments then use this state to show that a single code implementation supplies a $Z(\theta)$ that is reusable across output languages. These reusable abstractions support different realizations of the same computation in distinct contexts, while also allowing reuse of information acquired from one of those realizations.

We then propose a kind of structured data that would favor such reusable representations, and validate this prediction using a minimal toy task that displays the same behavioral signatures of compositionality that we observe in LLMs. We also show that accurate prediction does not guarantee an accessible, reusable suffix state, with a similarly performant toy model that uses dense attention only. It is worth noting, however, that these results do not establish definitively that the same data structure causes the analogous representations in LLMs.

Although we have studied code, little in our analysis depends on it. Programming is simply a domain that makes the shared structure easy to see and test, since one verifiable algorithm is written in many languages and one language carries many algorithms. \textbf{Much of the natural language corpus that LLMs learn from in pretraining also has the same shape}: a relatively small number of underlying mechanisms expressed and recombined into a vast number of distinct forms. The same pressure toward reusable, composable states should apply whenever that is true. From this perspective, \textbf{understanding where these shared representations form and how they compose is key to a robust understanding of how models organize what they know}. Our ongoing work aims to make the connection between shared mechanisms in the data and shared representations in the model increasingly precise, approaching a prediction of their exact geometries.

\section*{Acknowledgments}
Many discussions with wonderful colleagues were had in the process of writing up this work. We especially thank Eric Michaud and Loren Amdahl-Culleton for valuable feedback on the experiments and the writing.

\bibliography{iclr2026_conference}
\bibliographystyle{iclr2026_conference}

\newpage
\appendix
\section{Appendix}

\subsection{Notation and Shared Intervention Details}
\label{app:notation}

\begin{center}
\small
\begin{tabular}{p{0.25\linewidth}p{0.65\linewidth}}
    \toprule
    symbol & meaning \\
    \midrule
    $\theta,\theta^\star$ & task of the recipient prompt and target task of the edit \\
    $f,\lambda,s$ & function, language, and strategy values; $?$ marks a factor left unspecified \\
    $x=\tau_\theta\Vert\sigma$ & prompt: task-bearing description followed by the shared suffix \\
    $Z,\widetilde Z,\overline Z$ & suffix state of one prompt, its edited state, and the mean state over prompts with given factor values \\
    $\Delta_F,\Delta_L,\Delta_S$ & edits of function, language, and strategy; $\alpha$ their dose \\
    $\Lambda$ & languages whose prompts serve as donors for a function edit \\
    $\mathcal L,\mathcal S,\mathcal D$ & edited layers, edited suffix positions, and masked description positions \\
    \bottomrule
\end{tabular}
\end{center}

\paragraph{Prompts and editable state.}
In a code prompt, the ``task-bearing prefix'' $\tau_\theta$ includes the fixed language template, padding that aligns suffix positions,
and the function/strategy specification. A specification can be given by a description or, in selected
experiments, a canonical function name. The toy's $\tau_\theta$ excludes the beginning-of-sequence token and
contains the observed tokens before the shared suffix. The positions selected for access removal,
$\mathcal D$, need not cover this entire
prefix. The chosen layers $\mathcal L$ and suffix positions $\mathcal S$ define the editable cache block, and every intervention adds a scaled displacement to it:
\begin{equation}
    Z=\{K_{\ell,i},V_{\ell,i}\}_{\ell\in\mathcal L,\,i\in\mathcal S},
    \qquad
    \widetilde Z=Z+\alpha\Delta.
    \label{appeq:kv-steering}
\end{equation}

Selected recurrent entries are included where the model's protocol uses
them. The rest of the recipient's persistent state is fixed during the edit. We use $Z$ without a
task subscript within a trial; $Z(f,\lambda)$ or $Z(f,\lambda,s)$ is shorthand when comparing
conditions. The notation ``$?$'' for an ambiguous recipient names a prompt construction, it does not refer to a literal token, a zero value activation vector,
nor a guaranteed uniform model belief.

\paragraph{Predictions after an intervention.}
Let $\theta$ denote the recipient's task condition and $\theta^\star$ its target condition, including
both the factors to change and those to preserve. For each trial,
$q^{\theta\to\theta^\star}(\xi\mid v)$ denotes the probability assigned to the next token
$\xi$ after the prescribed intervention and subsequent processing of the continuation prefix $v$.
The recipient prompt and its unedited state entries, the donor estimate, the edit dose, and the
description-access rule are specified by the trial protocol; the arrow labels the intended change and does
not assert that it succeeded. When the intended change is clear, $q^{\mathrm{edit}}$ denotes the same
predictor, and a dot ``$\cdot$'' in place of $\xi$ denotes the whole next-token distribution. The prefix $v$ is the
generated continuation so far in code decoding or a supplied diagnostic prefix in the toy; an empty $v$
denotes the readout at the suffix boundary. $q(\xi\mid h)$ in the optimal-loss argument
denotes ordinary prediction from the full observed history $h$, without a state intervention. A
completion $u$ is the decoded program in the code experiments; $Q$ collects the toy's readout
distributions, the language distribution at the last blank and the distributions after the three probes.
The target enters their evaluation through the unit tests 
\begin{equation}
    \operatorname{Pass}_{\theta^\star}(u)\quad\text{for code},
    \qquad
    \operatorname{Pass}_{\theta^\star}(Q)\quad\text{for the toy}.
    \label{eq:intervention-score}
\end{equation}
These functions pass only when the target's task tests and all required factor readouts pass; reported
scores average it over trials.

\paragraph{Additive mean states.}
For function and language, the additive hypothesis applied to mean states is
\begin{equation}
    \overline Z(f,\lambda)\approx Z_0+Z_F(f)+Z_L(\lambda).
    \label{eq:toy-direct-sum}
\end{equation}
The same idealization extends to prompts with a factor unspecified, which is then denoted with ``$?$''. The recipient contributes its
own measured state $Z$; the displacement is estimated from donor means.

\paragraph{Mean states and held-out contexts.}
For a selected factor $j\in\{F,L,S\}$, let $a$ and $b$ be two of its values and $\gamma$ the remaining task
context, that is, the values of the other factors. Let $Z^{(i)}(a,\gamma)$ be the suffix state of the $i$-th
of $N$ donor prompts with value $a$ in context $\gamma$, and $\overline Z(a,\gamma)$ the mean of these states.
The displacement for the factor averages the difference of means over a set of contexts $\Gamma_j$:
\begin{align}
    \overline Z(a,\gamma)&=\frac{1}{N}\sum_{i=1}^{N}Z^{(i)}(a,\gamma),\nonumber\\
    \Delta_j[a\leftarrow b]&=\frac{1}{|\Gamma_j|}
        \sum_{\gamma\in\Gamma_j}
        \left[\overline Z(a,\gamma)-\overline Z(b,\gamma)\right].
    \label{eq:general-donor-estimator}
\end{align}
In the minimal process, paired prompts share blank positions and choices of shared versus language-specific
tokens. A language contrast keeps the function subtokens fixed; a function contrast uses independently
sampled function sequences. In language models, we replace the relevant specification and adjust padding to
keep the suffix tokens and their positions matched.

Pairing is not needed to estimate a difference of population means, but can reduce sampling noise.
Under the additive hypothesis, with other prompt variation controlled, context contributions approximately
cancel in the contrast. Averaging alone does not remove interactions between factors.
For leave-one-context-out estimation, the recipient's context
$\gamma^\star$ is excluded from $\Gamma_j$; the same-context-only comparison uses $\Gamma_j=\{\gamma^\star\}$.
Installation uses the same estimator with $b=?$, with the construction of the unspecified prompt given
separately for each setting. Composition adds two separately estimated installation displacements.

\paragraph{Interventions.}
The three tests in Section~\ref{sec:shared-interventions} specialize Eq.~(\ref{appeq:kv-steering}) to
$\alpha=1$ (See Appendix~\ref{app:dosage} for more dose testing). 
Function \emph{transport} starts from $(f,\lambda)$ and targets $(f^\star,\lambda)$:
\begin{equation}
    \widetilde Z=Z+\Delta_F[f^\star\leftarrow f].
    \label{eq:transport}
\end{equation}
Function \emph{installation} instead starts from the prompt $(?,\lambda)$, with the same target:
\begin{equation}
    \widetilde Z=Z+\Delta_F[f^\star\leftarrow ?].
    \label{eq:installation}
\end{equation}
\emph{Composition} starts from $(?,?)$ and targets $(f^\star,\lambda^\star)$:
\begin{equation}
    \widetilde Z=Z+\Delta_F[f^\star\leftarrow ?]+\Delta_L[\lambda^\star\leftarrow ?],
    \label{eq:composition}
\end{equation}
Language and strategy edits follow the same construction. The same-context-only estimate is a useful empirical comparison, but does not serve as an upper bound on performance.

\subsection{Shared-Mechanism Coordinates}
\label{app:shared-mechanism}

In principle, token sequences for any stochastic process can always be generated via some linear dynamic over a sufficiently large latent space.

Quantum mechanics exemplifies how this plays out in the physical world: 
our complex and apparently nonlinear world arises via linear evolution of its underlying quantum state.\footnote{In fact, linearity is a bit of a misnomer here.  The joint unitary evolution operator is linear, and the superoperator associated with a (POVM) observation is linear, but renormalizing a density matrix \emph{given} the observation induces a slight nonlinearity which technically induces a \emph{perspective transformation}.  In an exactly analogous way, we talk about the linear transition operator $T^{(x)}$, although the net transformation in the predictive-state update is a perspective transformation once we condition on a particular observation.}  There is an analogous mathematical treatment that is always valid as a probabilistic computational paradigm.
In assuming this mathematical framework, we are building on the recent approach and interpretability insights of 
\cite{shai2024transformers}, \cite{riechers2025neural}, \cite{factoredworld2026}, and \cite{ray2026nonergodic}.

In this framework, the joint probability of any token sequence can be expressed as
$\Pr(X_{1:L}) = \boldsymbol{\eta}^{(\varnothing)} T^{(X_{1:L})} \boldsymbol{1}$
where 
$T^{(X_{1:L})} = T^{(X_{1})} \cdots T^{(X_{L})}$ is the matrix multiplication of latent transition operators $T^{(x)}$, while 
$\boldsymbol{\eta}^{(\varnothing)}$ is the initial vector in the latent space at the beginning of context, and $\boldsymbol{1}$ is a fixed linear functional on the latent space that effectively integrates latent probability. For an HMM, as in the toy model we describe below, this is simply an all-ones vector with the dimension of the latent space.

In the present work, we explore the implications of further structure in the process:
specifically the case of a mixture of many different scenarios, where each scenario integrates contributions from many parts, and some of the parts may be shared across the scenarios.

 Let $T^{(\xi)}$ be the mixture's linear evolution operator which accounts for both state evolution and the probability of emitting $\xi$. Take $T_{\mathrm{shared}}^{(\xi)}$ to be the shared mechanisms operator with $T_\kappa^{(\xi)}$ collecting the unshared dynamics of the component $\kappa$, which may itself have multiple factors.
In the direct-sum form, states are grouped first by mixture component,
with the shared and remaining state coordinates inside each component. Reordering them to place the
shared coordinate first gives the equivalent representation
\begin{equation}
    T^{(\xi)}
    =\bigoplus_\kappa T_{\mathrm{shared}}^{(\xi)}\otimes T_\kappa^{(\xi)}
    =P\left(T_{\mathrm{shared}}^{(\xi)}\otimes
        \bigoplus_\kappa T_\kappa^{(\xi)}\right)P^{-1}.
    \label{eq:direct_sum_map}
\end{equation}
The permutation matrix $P$ here only
reorders the joint-state coordinates and is the same for every emitted token $\xi$. Thus the two forms describe identical dynamics, with the shared mechanism written once in the factored form.

Invoking the FWH \citep{factoredworld2026} suggests two different representations with distinct dimensional requirements: the former induces a transformation on 
the vector space $\mathcal{V}_\text{redundant} := \bigoplus_\kappa (\mathcal{V}_{\mathrm{shared}} \oplus \mathcal{V}_\kappa)$,
whereas the latter induces a transformation on the vector space 
$\mathcal{V}_\text{abstract} := \mathcal{V}_{\mathrm{shared}} \oplus
        \left(\bigoplus_\kappa \mathcal{V}_\kappa\right)$. The latter representation has effectively identified an abstraction---that the first part is shared among all scenarios---and represents this part only once, whereas the redundant representation would allocate a new space per scenario for the same part.
For example, if $\text{dim}(\mathcal{V}_{\mathrm{shared}}) = 10$ and $\text{dim}(\mathcal{V}_\kappa) = 5$ for all scenarios $\kappa$, and there are say $\kappa =20$ scenarios,
then 
$\text{dim}(\mathcal{V}_\text{redundant}) = \sum_{\kappa=1}^{20} (10+5) = 20 \times 15 = 300$
while 
$\text{dim}(\mathcal{V}_\text{abstract}) = 10+ \sum_{\kappa=1}^{20} 5 = 10+20 \times 5 = 110$.
The abstracted option is much more dimensionally efficient! Our toy model below factors in much the same way, with the abstract representation taking up $\approx20$ dimensions and the  redundant one $\approx 200$.

\subsection{The Minimal Programming Process}
\label{app:toy-process}

In our minimal model, 
each sequence uses one language $\lambda$ and one function $f$, drawn independently and uniformly from
eight languages and three functions. These choices stay fixed throughout the sequence. The data generating process
can be thought of as a single large hidden Markov model (HMM), or as a composition of a language machine, a function machine, and a shared ``watcher'' communicating through subtokens. Below we describe the process in terms of these interacting machines.  

\paragraph{Conventions.}
Each machine can be described by a set of linear operators on some latent subspace, as described in Appendix~\ref{app:shared-mechanism}. For each language--function pair $(\lambda, f)$, we index
the three co-active machines by $M\in\{\lambda,f,W\}$; each has a finite state set $\mathcal Q_M$ which can be thought of as a basis for the latent subspace, an initial vector
$\boldsymbol\eta^{(\varnothing)}_M$ in this subspace, and one transition operator per subtoken configuration.
The matrix element $T^{(a)}_{s,s'}$ is the probability of the transition $s\to s'$ while emitting subtoken $a$. Because the machines are interacting, the transition operators for one are often conditioned on subtokens that another machine emits. To denote this, we write the subtoken the machine emits and, after a vertical bar $|$, the subtokens of the other machines on which the operator is conditioned: $T_{\lambda}^{(c)}$,
$T_{f}^{(k\mid r)}$, and $T_W^{(r\mid c,k)}$ (see Table~\ref{tab:toy-machines}). In the following, $I_M$ is the identity on
$\mathcal Q_M$ and $\boldsymbol{1}$ the all-ones column vector.

\begin{table}[H]
\centering\small
\begin{tabular}{llllll}
\toprule
machine & state & $\mathcal Q_M$ & $\boldsymbol\eta^{(\varnothing)}_M$ & emits & conditions on \\
\midrule
language $\lambda$ & $z$ & $\{C,F\}$ & $(1,0)$ & $c\in\{S,L_\lambda,\square\}$ & -- \\
function $f$ & $g$ & $\mathcal Q_f$ & $\boldsymbol\eta^{(\varnothing)}_{f}$ & $k\in\{0,1,2,R,\square\}$ & $r$ \\
watcher $W$ & $m$ & $\{\emptyset,0,01\}$ & $(1,0,0)$ & $r\in\{\mathrm{cont},R,\square\}$ & $c,k$ \\
\bottomrule
\end{tabular}
\caption{The three machines in a common format: state variable, state set, initial vector, the
subtoken each emits, the subtokens of the other machines on which its
operator is conditioned. $\mathcal Q_f$ and $\boldsymbol\eta^{(\varnothing)}_{f}$ depend on the function and are given at the end of this
section.}
\label{tab:toy-machines}
\end{table}

\paragraph{Subtokens and emission order.}

\begin{figure}[H]
    \centering
    \includegraphics[width=1.0\linewidth]{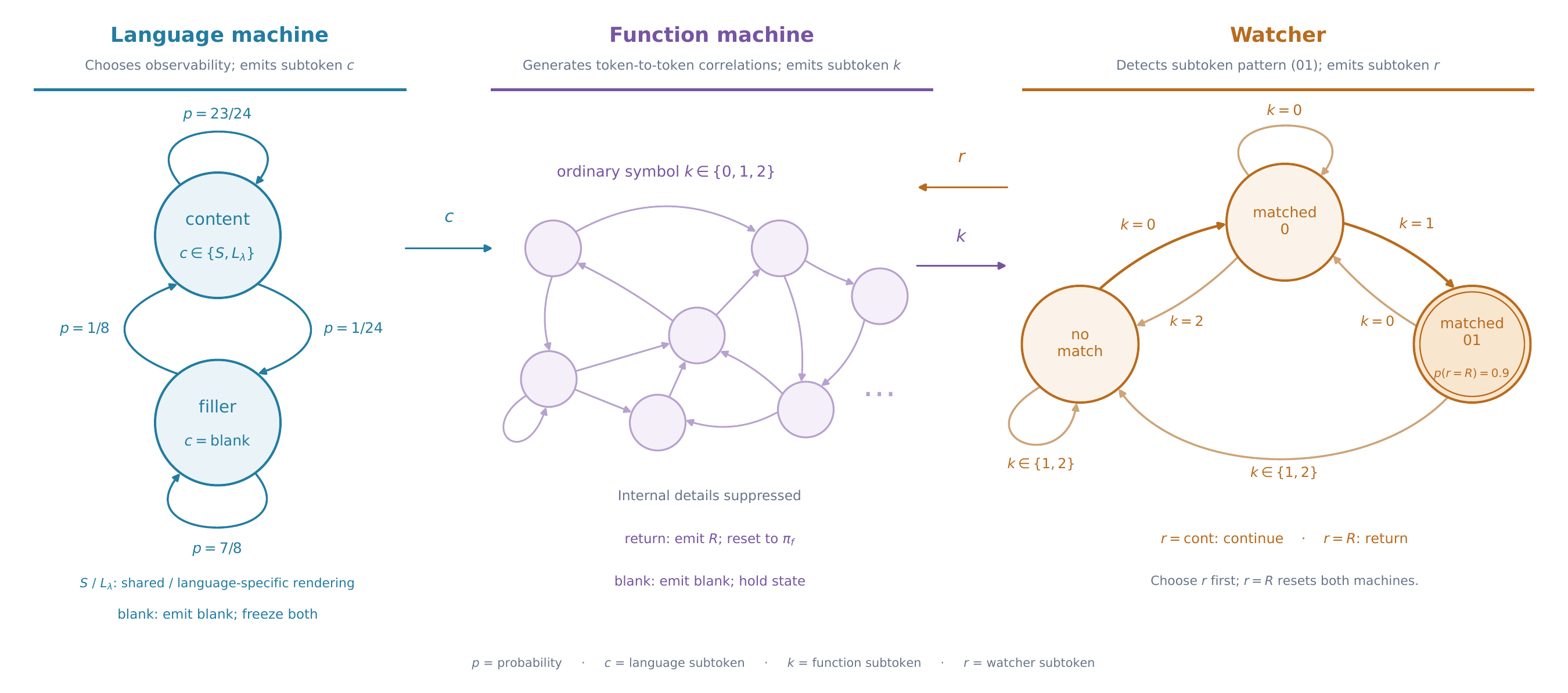}
    \caption{\textbf{Three coupled machines generate the observed tokens.} The function emits
    subtokens, the language selects shared or language-specific tokens and blank intervals, and the watcher
    triggers returns with a probability that rises after a pattern match. A return resets the function
    and watcher; blanks preserve the function and watcher states. The machines communicate through internal subtokens,
    combined by a deterministic map into the observed token stream. Function states in the diagram are
    schematic.}
    \label{fig:toy-process}
\end{figure}
Each machine emits a subtoken, with an alphabet defined for its role:
\[
    c \in
    \{S,L_\lambda,\square\},
    \qquad
    k\in\{0,1,2,R,\square\},\qquad
    r\in\{\mathrm{cont},R,\square\}.
\]
The language subtoken $c$ selects whether we see a shared token ($S$), a language-specific token ($L_\lambda$), or a blank
($\square$). The watcher subtoken $r$ tells the function to continue, return, or hold. The function
subtoken $k$ supplies an ordinary symbol when told to continue, and otherwise mirrors the watcher's return and hold instructions. One step of the full process proceeds as follows:
\begin{enumerate}
    \item \textbf{Language-machine update.} 
    The language machine emits $c$ from its current mode. If $c=\square$, both other machines
    emit $\square$ and hold their states. The language mode can still change.
    \item \textbf{Watcher subtoken without transition.}
    On a content step, $c\in\{S,L_\lambda\}$, the watcher chooses $r$ using its current state without any latent transition.
    \item 
    \textbf{Function-machine update.}
    If $r=\mathrm{cont}$, the function emits $k\in\{0,1,2\}$ and updates its state using function-specific operators. 
    If $r=R$, the function also emits $k=R$ and resets to $\boldsymbol\eta^{(\varnothing)}_{f}$. %
    \item 
    \textbf{Watcher-subspace update.} The watcher
    then updates its state using $k$. If $k=R$, the watcher resets to $\emptyset$.
\end{enumerate}

The first three of these substeps can be implemented by the linear map 
\begin{align}
\bigoplus_{\lambda, f} T_\lambda^{(c)} \otimes D_W^{(r|c)} \otimes T_f^{(k|r,c)}~, 
\end{align}
where 
we've defined the conditional diagonal matrix 
$D_W^{(r|c)} := \sum_{m} \Pr(r|c,m) |m \rangle \! \langle m |$ that acts on the latent watcher subspace.
Here, bras and kets denote row vectors and column vectors, respectively.

The fourth substep, 
where we apply a transition on the latent watcher subspace, conditioned on k, can then be implemented by matrix multiplication from the right by 
\begin{align}
\bigoplus_{\lambda, f}  I_\lambda \otimes T_W^{|c,k} \otimes I_f ~,
\end{align}
where 
$T_W^{|c,k} := \sum_{m,m'} \Pr(m'|c,k,m) | m \rangle \! \langle m' |$,
yielding the composite token-labeled transition
\begin{align}
T^{(\xi)} = \bigoplus_{\lambda, f} T_\lambda^{(c)} \otimes T_W^{(r\mid c,k)} \otimes T_f^{(k|r,c)} ~,
\label{eq:JointToyTransition}
\end{align}
where $T_W^{(r\mid c,k)} = D_W^{(r|c)} T_W^{|c,k}$.

\paragraph{Observed tokens.}
Based on the emitted subtokens, the process emits only one observable token $\xi=\phi(c,k,r)$ per step, according to
\begin{equation}
    \phi(c,k,\mathrm{cont})=c_k,\qquad
    \phi(\square,\square,\square)=\square,\qquad
    \phi(c,R,R)=c_R,
    \label{eq:toy-observation-map}
\end{equation}
where $c\in\{S,L_\lambda\}$ for content and $k\in\{0,1,2\}$ for ordinary symbols. If we now treat $S$ and $L_\lambda$ as sets, the labels $c_k$ and
$c_R$ denote shared or language-specific tokens in $S \cup L_\lambda$, for symbol $k$ and return $R$. Only these triples occur,
so each observed token identifies a unique subtoken triple. An equivalent single HMM emits $\xi$
directly according to Eq.~(\ref{eq:JointToyTransition}). The observed vocabulary across all languages is
\[
    \{\square,S_0,S_1,S_2,S_R\}
    \cup\{L_{\lambda,0},L_{\lambda,1},L_{\lambda,2},L_{\lambda,R}\}_{\lambda=0}^7.
\]
It contains $37$ tokens. A prepended beginning-of-sequence token gives the model $38$ vocabulary entries.

\paragraph{Language machine.}
States $z\in\{C,F\}$ (denoting ``content" or ``filler" modes), initialized to $\boldsymbol\eta^{(\varnothing)}_{\lambda}=(1,0)$. The
emission depends only on the current mode: in $C$ the machine emits $S$ with probability $0.8$ and
$L_\lambda$ with probability $0.2$; in $F$ it emits a blank. The mode then changes independently of what
was emitted, $C\to F$ with probability $\Delta$ and $F\to C$ with probability $\Gamma$, so content and filler
runs have mean lengths $\Delta^{-1}$ and $\Gamma^{-1}$, respectively. With modes ordered as $(C,F)$,
\begin{equation}
    T_{\lambda}^{(S)}=0.8\begin{pmatrix}1-\Delta&\Delta\\0&0\end{pmatrix},\qquad
    T_{\lambda}^{(L_\lambda)}=0.2\begin{pmatrix}1-\Delta&\Delta\\0&0\end{pmatrix},\qquad
    T_{\lambda}^{(\square)}=\begin{pmatrix}0&0\\1-\Gamma&\Gamma\end{pmatrix},
    \label{eq:toy-language-operators}
\end{equation}
and $T_{\lambda}^{(L_{\lambda'})}=0$ for $\lambda'\neq\lambda$. All langugaes share the same preference for content over filler with $\Delta^{-1}= 24$ and $\Gamma^{-1}=8$~.

\paragraph{Function machine.}
States $g\in\mathcal Q_f$ and start $\boldsymbol\eta^{(\varnothing)}_{f}$ are function-specific and are given at the end of this
section. The function operators are conditioned only on the watcher subtoken $r$, which selects one of three
operators: normal content, a reset, or a blank step:
\begin{equation}
    T_{F_f}^{(k\mid r)}=
    \begin{cases}
        A_f^{(k)},&r=\mathrm{cont},\ k\in\{0,1,2\},\\
        R_f=\boldsymbol{1}\,\boldsymbol\eta^{(\varnothing)}_{f},&r=k=R,\\
        I_F,&r=k=\square,\\
        0,&\text{otherwise}.
    \end{cases}
    \label{eq:toy-function-operators}
\end{equation}
The content operators $A_f^{(k)}$ are what distinguish the three
functions. While the different functions provide different token statistics, every ordinary symbol has positive probability from
every state of every function, so no function is ever ruled out by a finite sequence of tokens. The reset $R_f$ has every row equal to $\boldsymbol\eta^{(\varnothing)}_{f}$, so a return sends the function to its initial
vector regardless of its current state. 

\paragraph{Watcher.}
States $m\in\{\emptyset,0,01\}$, initialized to $\boldsymbol\eta^{(\varnothing)}_{W}=(1,0,0)$. The watcher tracks
whether the function's recent content symbols end in $0\,1$ and, on each content step, triggers a return (subtoken $R$) with
probability $\alpha$ if they do (returns can be spontaneously triggered otherwise, with a probability of $\epsilon$).
On a blank step, where $c = \square$, the function and watcher both emit a blank too without any latent transition: $T_W^{(\square\mid\square,\square)}=I_W$.
For all other $c \neq \square$, and
with states ordered as $(\emptyset,0,01)$,
the conditional substochastic transition matrices are
\begin{align}
    T_W^{(R\mid c)}&=\begin{pmatrix}\epsilon&0&0\\\epsilon&0&0\\\alpha&0&0\end{pmatrix},&
    T_W^{(\mathrm{cont}\mid c,0)}&=\begin{pmatrix}0&1-\epsilon&0\\0&1-\epsilon&0\\0&1-\alpha&0\end{pmatrix},\nonumber\\
    T_W^{(\mathrm{cont}\mid c,1)}&=\begin{pmatrix}1-\epsilon&0&0\\0&0&1-\epsilon\\1-\alpha&0&0\end{pmatrix},&
    T_W^{(\mathrm{cont}\mid c,2)}&=\begin{pmatrix}1-\epsilon&0&0\\1-\epsilon&0&0\\1-\alpha&0&0\end{pmatrix}.
    \label{eq:toy-watcher-operators}
\end{align}
Our watchers all use $\alpha=.9$ and $\epsilon=.01$~.

\paragraph{Joint source.}
For each observed token, the three operators are evaluated at the subtoken triple and combined by a
tensor product; the direct sum runs over the fixed language and function choices:
\begin{equation}
    T^{(\xi)}=\bigoplus_{\lambda,f}
    T_{\lambda}^{(c)}\otimes T_{f}^{(k\mid r)}\otimes T_W^{(r\mid c,k)},
    \qquad (c,k,r)=\phi^{-1}(\xi).
    \label{eq:toy-source}
\end{equation}
With initial vector
$\boldsymbol\eta^{(\varnothing)}=\tfrac1{24}\bigoplus_{\lambda,f}\boldsymbol\eta^{(\varnothing)}_{\lambda}\otimes\boldsymbol\eta^{(\varnothing)}_{f}\otimes\boldsymbol\eta^{(\varnothing)}_{W}$ , we have
$\Pr(\xi_{1:L})=\boldsymbol\eta^{(\varnothing)}T^{(\xi_1)}\cdots T^{(\xi_L)}\boldsymbol{1}$ as in
Appendix~\ref{app:shared-mechanism}. To see Eq.~(\ref{eq:direct_sum_map}) in this source, hold the function fixed. The eight languages are then
the scenarios $\kappa=\lambda$, the function machine together with the watcher is the mechanism they
share, and the $f$-th block of Eq.~(\ref{eq:toy-source}) is
\begin{align}
    \bigoplus_\lambda T_{\lambda}^{(c)}\otimes T_{f}^{(k\mid r)}\otimes T_W^{(r\mid c,k)}
    &=P\Bigl[\bigl(T_{f}^{(k\mid r)}\otimes T_W^{(r\mid c,k)}\bigr)\otimes
        \bigoplus_\lambda T_{\lambda}^{(c)}\Bigr]P^{-1},
    \label{eq:toy-shared-first}
\end{align}
which is Eq.~(\ref{eq:direct_sum_map}) with $T_{\mathrm{shared}}=T_{f}\otimes T_W$ and
$T_\kappa=T_{\lambda}$. The full source is the direct sum of these blocks over $f$. Holding the language
fixed instead makes the three functions the scenarios and the language machine together with the watcher
the shared mechanism. In either case $P$ only reorders tensor factors and is the same for every $\xi$.

Note that, given $\lambda$ and $f$, the likelihood of a sequence factors into a language term, a function term, and a
watcher term that takes the same form for every $(\lambda,f)$, because the watcher's state is a deterministic
function of the observed subtokens. Independent priors over $\lambda$ and $f$ therefore stay a product ($\eta_\lambda \otimes\eta_f$) when being updated into the posterior, despite the coupling of the
function and watcher dynamics.

\paragraph{The function operators.}
Each function is specified by its content operators $A_f^{(k)}$; a return
(Eq.~\ref{eq:toy-function-operators}) sends it back to its initial vector, the point mass on its start
state unless stated otherwise.

\begin{itemize}[leftmargin=*]
    \item \textbf{Mess3} drifts slowly among three states, each favoring its own symbol. Its start is
    uniform, $\boldsymbol\eta^{(\varnothing)}_{f}=(\tfrac13,\tfrac13,\tfrac13)$, and with states ordered
    $(0,1,2)$,
    \[
        {\setlength{\arraycolsep}{3pt}
        A_f^{(0)}=\begin{pmatrix}0.42&0.03&0.03\\0.09&0.14&0.03\\0.09&0.03&0.14\end{pmatrix},\quad
        A_f^{(1)}=\begin{pmatrix}0.14&0.09&0.03\\0.03&0.42&0.03\\0.03&0.09&0.14\end{pmatrix},\quad
        A_f^{(2)}=\begin{pmatrix}0.14&0.03&0.09\\0.03&0.14&0.09\\0.03&0.03&0.42\end{pmatrix}.}
    \]
    \item \textbf{No-repeat} only emits the same symbol twice in a row with probability $\epsilon$. Otherwise it selects from the other two symbols with equal probability $\gamma=(1-\epsilon)/2$  It has a transient
    start state form which is selects uniformly, and states $\mathsf{last}_g$ for $g\in\{0,1,2\}$ (which records the previous symbol); with states
    ordered $(\mathsf{start},\mathsf{last}_0,\mathsf{last}_1,\mathsf{last}_2)$,
    {\setlength{\arraycolsep}{3pt}
    \begin{align*}
        A_f^{(0)}&=\begin{pmatrix}0&1/3&0&0\\0&\epsilon&0&0\\0&\gamma&0&0\\0&\gamma&0&0\end{pmatrix},&
        A_f^{(1)}&=\begin{pmatrix}0&0&1/3&0\\0&0&\gamma&0\\0&0&\epsilon&0\\0&0&\gamma&0\end{pmatrix},\\
        A_f^{(2)}&=\begin{pmatrix}0&0&0&1/3\\0&0&0&\gamma\\0&0&0&\gamma\\0&0&0&\epsilon\end{pmatrix}.
    \end{align*}}
    Our function machines use $\epsilon=1/20$.
    \item \textbf{Noisy sum modulo three} emits triples $a$, $b$, $(a+b)\bmod 3$, but has a probability of $\gamma$ to select each of the wrong answers on the third step. Its states are: the start state; $\mathsf{got}_a$ for
    $a\in\{0,1,2\}$, recording the first symbol of each triple; and $\mathsf{need}_j$ for $j\in\{0,1,2\}$,
    recording the correct sum to emit; with states ordered
    $(\mathsf{start},\mathsf{got}_0,\mathsf{got}_1,\mathsf{got}_2,\mathsf{need}_0,\mathsf{need}_1,\mathsf{need}_2)$,
    {\setlength{\arraycolsep}{2pt}
    \begin{align*}
        A_f^{(0)}&=\begin{pmatrix}0&1/3&0&0&0&0&0\\0&0&0&0&1/3&0&0\\0&0&0&0&0&1/3&0\\0&0&0&0&0&0&1/3\\\epsilon&0&0&0&0&0&0\\\gamma&0&0&0&0&0&0\\\gamma&0&0&0&0&0&0\end{pmatrix},&
        A_f^{(1)}&=\begin{pmatrix}0&0&1/3&0&0&0&0\\0&0&0&0&0&1/3&0\\0&0&0&0&0&0&1/3\\0&0&0&0&1/3&0&0\\\gamma&0&0&0&0&0&0\\\epsilon&0&0&0&0&0&0\\\gamma&0&0&0&0&0&0\end{pmatrix},\\
        A_f^{(2)}&=\begin{pmatrix}0&0&0&1/3&0&0&0\\0&0&0&0&0&0&1/3\\0&0&0&0&1/3&0&0\\0&0&0&0&0&1/3&0\\\gamma&0&0&0&0&0&0\\\gamma&0&0&0&0&0&0\\\epsilon&0&0&0&0&0&0\end{pmatrix}.
    \end{align*}}
\end{itemize}
We set $\epsilon=1/20$ for our toy process.
In this case, it may help intuition to note that these function operators can be written as 
\begin{align*}
A_f^{(k)}=\begin{pmatrix}
            \mathbf 0 & \tfrac13\,\mathbf e_k & \mathbf 0_{1\times 3}\\
            \mathbf 0_{3\times 1} & \mathbf 0_{3\times 3} & \tfrac13\,\Pi^k\\
            \mathbf v^{(k)} & \mathbf 0_{3\times 3} & \mathbf 0_{3\times 3}
        \end{pmatrix}~,
\end{align*}
where 
$\Pi := \begin{pmatrix}
            0 & 1 & 0 \\
            0 & 0 & 1 \\
            1 & 0 & 0 
        \end{pmatrix}$
is a simple permutation matrix, 
$\mathbf e_k $ is a one-hot row vector,
and $\mathbf v^{(k)} := \Pi^{-k} \begin{pmatrix}
            \epsilon \\
            \gamma \\
            \gamma  
        \end{pmatrix} $ 
and note that $\Pi^{-1} = \Pi^\top$.

\subsubsection{Toy model results}
\label{app:toy-outside}
\begin{figure}[H]
    \centering
    \includegraphics[width=0.85\linewidth]{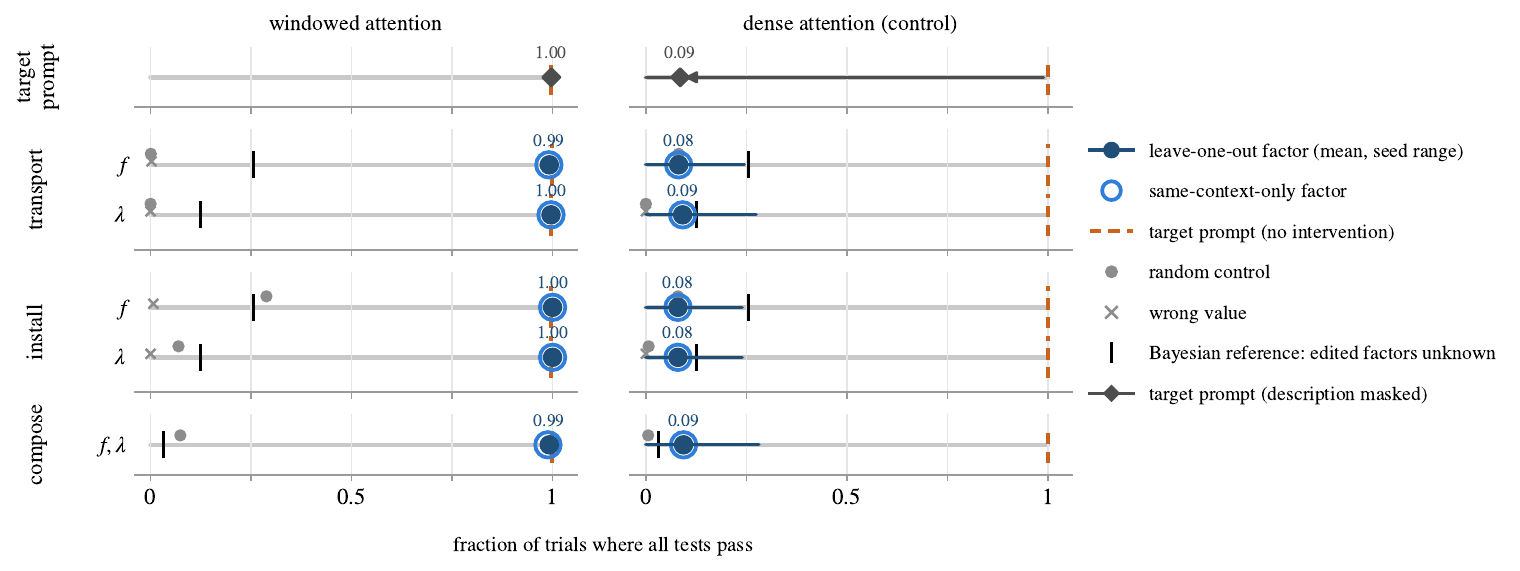}
    \caption{\textbf{Function and language support causal reuse in the toy model's suffix state.}
    Three training seeds are pooled (marker: mean; bar:
        seed range). Each row shows the target prompt performance (dashed orange); The top line of each column shows target prompt with its description masked (gray diamond, averaged within the columns because the drop is about the same on every row). Filled blue: leave-one-context-out edit; open blue: same-context-only edit; gray: random-direction and wrong-value controls. Black tick: Bayesian reference with uniform priors over unknown factors and the remaining factors known. We compare with explicit prompting of the target task, the target task prompt under the same mask, and the unedited, masked recipient sequence. We also include a wrong-value control that adds an edit toward an incorrect factor value while retaining the original target test.}
    \label{fig:toy-cfs}
\end{figure}

Figure \ref{fig:toy-cfs} shows the transport, installation, and composition tests work nearly perfectly for all recipients and donors in the windowed attention model. We see that these interventions work equally well whether they use same-context-only or leave-one-out steering---supporting the claim that this toy model has learned composable, reusable representations for language and function.

The dense attention model achieves similarly low prediction loss but displays much weaker function reuse at the suffix positions. We note that the failure of these interventions merely shows that there is not a reusable representation \emph{at the shared suffix}---not that the dense models do not have a reusable representation. This is consistent with the model's poor performance under prompt masking, even when given the correct target prompt.

\begin{figure}[H]
    \centering
    \includegraphics[width=0.8\linewidth]{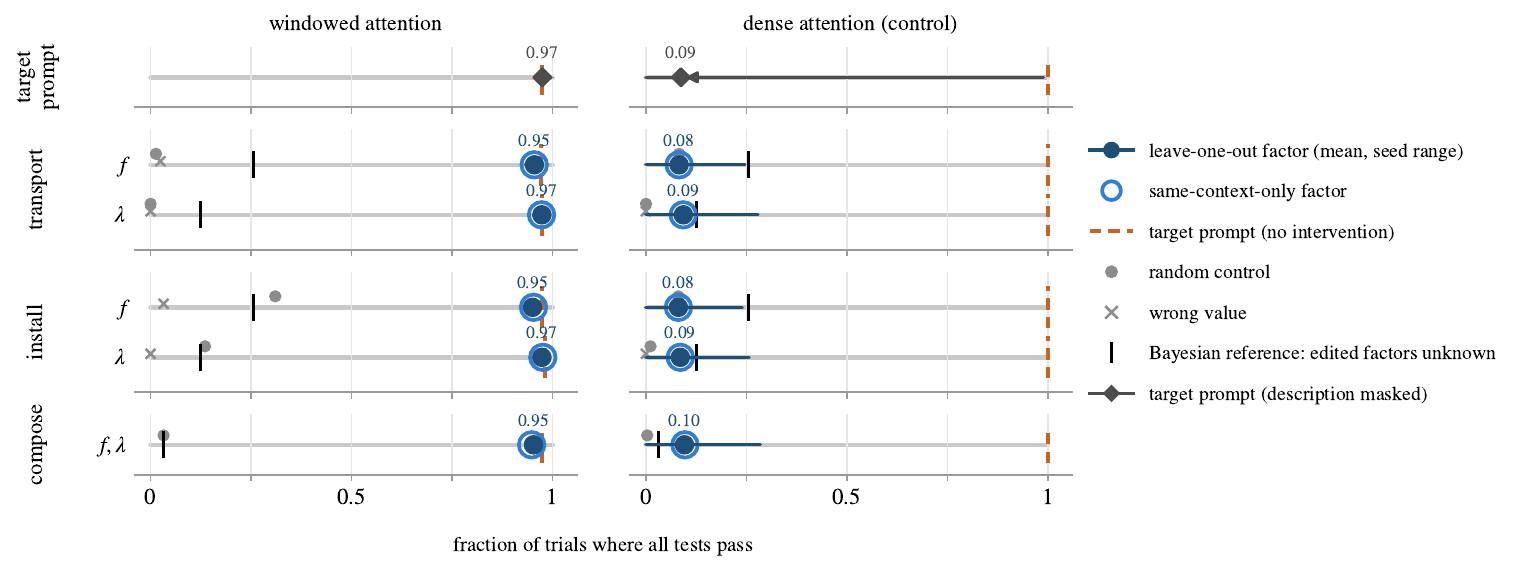}
    \caption{\textbf{With sixteen blanks the windowed model shows the same behavior without direct
    description access.} Same layout and conventions as Fig.~\ref{fig:toy-cfs}, including the Bayesian reference
    ticks. For the windowed model, masking the description is provably inert, so the diamond sits exactly on the dashed line.}
    \label{fig:toy-cfs-outside}
\end{figure}

Figure~\ref{fig:toy-cfs-outside} shows the same interventions performed with a suffix of 16 blank tokens instead of 4. Here, the masking of the prefix is vacuous, by construction because the model architecture cannot attend through the 16 blanks anyway due to an attention window of 8 tokens. In fact, not even the blanks that the model \emph{can} attend to have access to the prompt. The longer-gap setting forces direct prediction to use retained state, whereas at four blanks direct description access would still be available without masking. Regardless, the intervention and the prompted, masked prompted behavior is similar. This supports the idea that, even at 4 blanks, the model chooses to compile the prompt. Both settings support function reuse in the windowed model. The dense control fails nearly all tests in each case.

\subsubsection{Details of toy model Training and Interventions}
\label{app:toy-training-protocol}

The windowed model has $8$ attention-and-MLP transformer blocks, residual-stream dimension $d_\text{model}= 64$, $4$ heads per attention layer, a multilayer
perceptron of width $256$, LayerNorm normalization, and rotary positional embeddings. In every layer, attention was
restricted to the current plus the seven previous positions. The dense control has $4$ transformer blocks of the same
widths but attends over the whole context. Both are trained for $50{,}000$ steps at batch $512$ with AdamW \cite{kingma2014adam}, \cite{loshchilov2017decoupled} at
learning rate $10^{-3}$, $1{,}500$ warm-up steps and cosine decay, from three seeds each. Over the three seeds, the average evaluation
losses exceed the optimal Bayesian observer's loss on the same evaluation batch by $7.6\times 10^{-4}$ (windowed) and $8.4\times 10^{-4}$ nats (dense).

\paragraph{Prompt construction.} Each $\tau_\theta$ has $36$ positions sampled from one of the joint machines. For the corresponding prompts used to estimate edits, we hold fixed the blank positions and the positions of shared versus language-specific tokens, forcing the first token to identify the language unless we desire a language ambiguous prompt. We independently sample a subtoken sequence per function. Changing $f$ selects a different function subtoken sequence while keeping everything else the same; changing
$\lambda$ replaces only the language-specific tokens but keeps function subtokens fixed. Every prompt has the same suffix of $B$ blanks.

\paragraph{Test cases and recipients.} From a pool of $90$ matched sets of candidate prompts, we take the first $24$ for which each of the three function descriptions gives the observer at least $0.95$ posterior probability on its true function. This avoids ambiguous recipients for which an edit could outperform ordinary prompting simply by supplying more certain function information.

A prompt $(?,\lambda)$ keeps the blank positions and token choices but uses an intentionally function-ambiguous
sequence. At each ordinary-symbol step, we select the symbol that keeps the observer's function posterior closest to uniform. At the end of the 36 tokens, even an optimal observer would not be able to do better than a uniform guess over the functions. A prompt $(f,?)$ expresses the sampled function subtokens using only shared tokens, but using a sequence of function subtokens that would leave an observer $>95\%$ certain about which function was active; $(?,?)$ uses only shared tokens on the function ambiguous sequence.

\begin{figure}
    \centering
    \includegraphics[width=\linewidth]{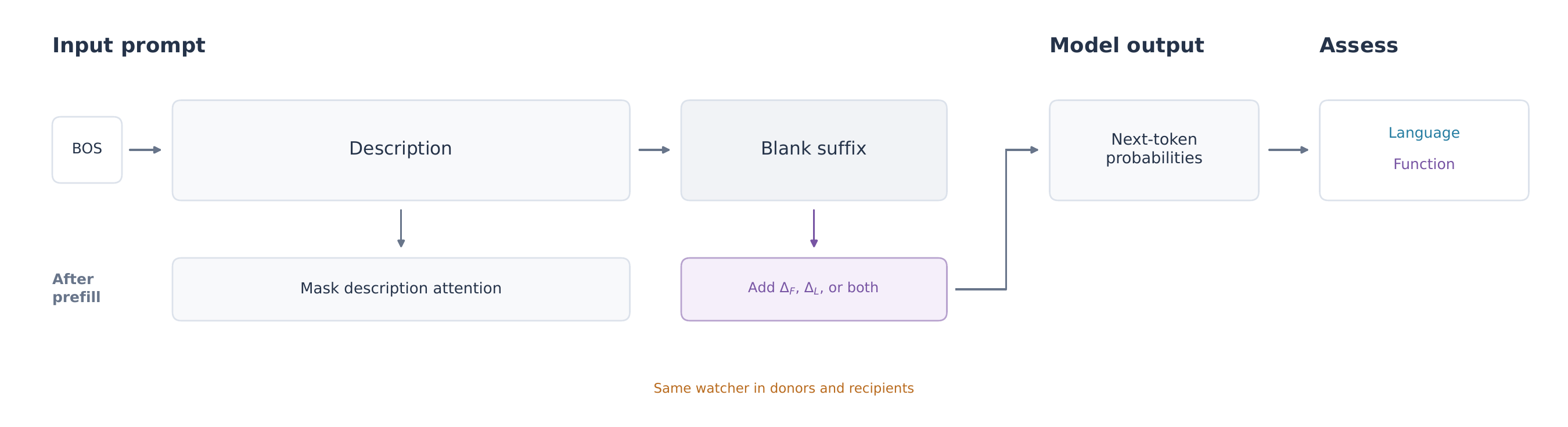}
    \caption{\textbf{Prompts, cache operations and readouts on the minimal process.} A prompt is a
    beginning-of-sequence token, a $36$-token description containing ordinary symbols, returns and
    blanks, and a suffix of $B$ shared blanks. After prefill the description's cached keys and
    values are masked out of attention and a function displacement $\Delta_F$, a language displacement $\Delta_L$ or
    their sum is added to the blank suffix's keys and values at every layer; the input tokens and the
    beginning-of-sequence cache do not change. The model then produces next-token distributions from the
    edited cache, and every test is a test on those distributions; no completion is sampled. The language is
    scored from the distribution immediately after the last blank. Each function test appends its own
    probe to a separate copy of the edited cache and scores the distribution after the probe.}
    \label{fig:toy-prompts}
\end{figure}

\paragraph{Masking the description.} After prefill, the $36$ description positions are excluded from attention in every layer (their scores set to $-\infty$ before the softmax) for every condition; the blanks' cache entries, written while the description was visible, are untouched. Only the readout queries are recomputed, so with a window of eight the ablation can matter only if a readout query has a content position in its window: the last blank sees content for $B\leq 7$ and the first probe
token for $B\leq 6$. For $B\geq 8$ the ablation changes nothing.

\paragraph{Correspondence with the code prompts.}
Both experiments use the prompt structure of Eq.~(\ref{eq:prompt-structure}) and the interventions of
Section~\ref{sec:shared-interventions}. Table~\ref{tab:toy-correspondence} summarizes their differences.
Masking the entire prompt in the toy model is necessary for the toy model to fully mask the toy model's ability to read function identity from $d_\theta$. In the natural language coding prompts, it is only necessary to mask the parts that specify the task. For example, ``Write a function'' need not be masked---but ``in python'' or ``that square's a number'' does.
\begin{table}[H]
    \centering
    \small
    \caption{Main differences between code prompts and the minimal process.}
    \label{tab:toy-correspondence}
    \begin{tabular}{p{0.18\linewidth}p{0.36\linewidth}p{0.36\linewidth}}
        \toprule
        element & code model & minimal process \\
        \midrule
        $\tau_\theta$ & instructions name the computation and language & $36$ tokens: statistics indicate function; tokens identify language \\ \\
        ? factors & omit the function or language instruction & use an ambiguous function sequence or only shared tokens, respectively \\ \\
        $\sigma$ & shared formatting instruction, with an empty think block & a string of shared blank tokens that hold function dynamics fixed \\ \\
        $\sigma$ alignment & pad $\tau_\theta$ to align the suffix & equal-length $\tau_\theta$ already matches blank positions \\ \\
        masking $\tau_\theta$ & positions naming the edited task  & all $36$ description positions \\ \\
        evaluation & execute the program and test the requested task & probe the predictive distributions for language tokens, and evaluate function on three probes \\ \\
        different contexts & ten languages, 31 tasks & eight languages, three tasks \\
        \bottomrule
    \end{tabular}
\end{table}

\paragraph{Edit vector Estimation.} We use $256$ prompts per language, function pair; this includes the prompts with either factor unspecified, using the procedure above. For each combination,
we average the keys and values associated with the suffix context positions across all $256$ prompts, separately at each position and layer. Displacements are differences of these means, added at the corresponding recipient suffix positions. Leave-one-context-out averages over the seven other languages for function edits and the two other functions for language edits; the same-context-only estimate uses the recipient's context only.

\paragraph{Controls} The random control draws a random direction norm-matched to the leave-one-out displacement. The wrong-value control uses a displacement toward a randomly chosen incorrect factor.

\paragraph{Readouts and the three probes.}
At the last blank, let $q_\lambda$ be the total probability assigned by $q^{\mathrm{edit}}$ to language
$\lambda$'s four private tokens, with an empty continuation after the shared prefix. The language readout passes when
the target has the largest $q_\lambda$ and the sum over all other languages is below $0.05$. Function scoring is trickier, because the observed tokens cannot be read directly for function identity. To query the model's function subtoken distribution $q^{\mathrm{sym}}$, we marginalize over the probabilities of all observed tokens that correspond to the same ordinary function subtoken, excluding blanks and returns. To probe for function identity, we insert a short probe sequence $v$ after the suffix, and measure $q^{\mathrm{sym}}(\cdot\mid v)$ for the specified probe. Each trial uses three probes from the family corresponding to its target function. Each $v$ starts with a return $R$ and uses only shared tokens. For the three functions, the probes follow the structure below.
\begin{center}
\begin{tabular}{p{0.23\linewidth}p{0.16\linewidth}p{0.52\linewidth}}
    \toprule
    target function & probe & pass rule \\
    \midrule
    sum modulo three & $R,a,b$ & probability of $(a+b)\bmod3$ exceeds $0.5$ \\
    no-repeat & $R,a$ & probability of $s$ is below $0.15$ \\
    Mess3 & $R,a,a,a$ & $D_{\mathrm{KL}}(q_{\mathrm{sym}}\Vert p_{\mathrm{Mess3}})$ is smaller than the divergence to
    either other function's predictive distribution \\
    \bottomrule
\end{tabular}
\end{center}
With $a,b$ sampled uniformly from $\{0,1,2\}$, independently across probes, and the same probes reused across conditions for that trial. An edit passes only if it passes three independently sampled function probes and the
language readout test.

\paragraph{Bayesian reference.} The black tick on each row of Figs.~\ref{fig:toy-cfs}
and~\ref{fig:toy-cfs-outside} is the expected score of a Bayes-optimal observer that knows everything about the trial except the edited factor, over which it holds a uniform prior. It sees the same probe tokens as the model, but it never sees the description. Its value depends only on the process and the probe. The function reference is {\toyRefF}, the language reference is the chance level {\toyRefL}, and the composition reference is their product, {\toyRefFL}. The function reference is not zero because the probe tokens are themselves evidence that favor some functions over others. This means an observer with no prior knowledge still passes some probes after updating on them. This reference measures how often our interventions would pass if the edits simply erased the model's evidence of the recipient task rather than creating certainty about the target task. This reference should not be read as a floor, since committing to the wrong function is worse than not knowing.

\paragraph{Estimating function beliefs.}
As mentioned above, the probe tokens are themselves evidence about the function: three repeated symbols are far more likely under Mess3 than under no-repeat. A model can therefore pass a probe without having retained the function, and this is why the unedited prompt and the random-direction control score {\tbAccWIFUned} and {\tbAccWIFCtrl} on function installation in Fig.~\ref{fig:toy-cfs} rather than zero. The pass rules for our function probes use the models predictive distribution \emph{after} seeing the probe, so they combine the models initial certainty about a function with the evidence the probe itself supplies. While the pass thresholds were designed to account for this, we include an additional measure here-- as supporting evidence that our probe functions are working as intended.

To separate retained information from evidence supplied by the probe, we now estimate the model's belief over functions directly. A model that became sure of the function only during the probe assigns lower probability to the symbols that convinced it, when compared to a model that was sure to begin. Thus, we can use these probabilities as a proxy for the models function beliefs. We append each of the possible 9 probes consisting of a return followed by function subtokens sampled from ${0,1,2}$; The probability of each continuation is the product of the model's next-symbol probabilities along it. The model's estimated belief is, then, the mixture of the three functions that gives a distribution over continuations that is closest to the model's (using Kullback--Leibler divergence as distance). This mixture can be used to quantify how strongly the model has committed to one function. We can easily optimize the mixture because each return token resets each function to a known state so we can assign a known probability distribution to each function and each three-symbol continuation exactly.

Table~\ref{tab:toy-belief} reports the estimated belief in the target function on the four-blank trials of Fig.~\ref{fig:toy-cfs}, rather than the percentage of trials that pass our set of tests. The results here are consistent with the results presented in the figure. While our function-specific probes \emph{could} have conflated uncertainty about the task with high performance, they do not in practice. These results support our suite of tests as well-calibrated. The belief estimating also makes it clear that the dense model looses all of the information in the prefix after masking: the dense attention control model is always uninformed after masking, and every edit leaves it similarly uncertain. ({\tbAccDTFLooFBel}).

\begin{table}[H]
    \centering
    \small
    \caption{Estimated belief in the target function on the four-blank trials of Fig.~\ref{fig:toy-cfs}, for the leave-one-context-out edit (mean over three training seeds, seed range in parentheses), the same-context-only edit, the unedited prompt, and the random-direction and wrong-value controls. A model certain of the target scores $1$, an uninformed one $1/3$, one committed to the wrong function $0$.
    Prompted models score {\tbAccWTFPromptFBel}.}
    \label{tab:toy-belief}
    {\setlength{\tabcolsep}{4pt}
\begin{tabular}{llccccc}
\toprule
model & row & \shortstack{leave-one-\\context-out} & \shortstack{same-\\context-only} & unedited & \shortstack{random\\direction} & \shortstack{wrong\\value} \\
\midrule
windowed & transport & 0.94 (0.94--0.94) & 0.94 & 0.01 & 0.01 & 0.01 \\
 & installation & 0.96 (0.95--0.96) & 0.96 & 0.33 & 0.33 & 0.02 \\
 & composition & 0.95 (0.95--0.96) & 0.96 & 0.33 & 0.34 & -- \\
\midrule
dense (control) & transport & 0.33 (0.33--0.34) & 0.33 & 0.33 & 0.33 & 0.33 \\
 & installation & 0.33 (0.33--0.34) & 0.33 & 0.33 & 0.33 & 0.33 \\
 & composition & 0.33 (0.33--0.34) & 0.33 & 0.33 & 0.33 & -- \\
\bottomrule
\end{tabular}}
 \end{table}

\subsection{Natural Language Prompting}
\label{app:prompting}

\paragraph{Prompt construction.}
The task-bearing prefix contains the specification, fixed template, padding that aligns suffix positions,
and beginning-of-sequence token. The suffix is a shared formatting instruction and, for some models,
an empty \verb|<think></think>| block. We generate without reasoning.

\paragraph{Persistent-state intervention.}
We apply Eq.~(\ref{appeq:kv-steering}) once to the suffix cache after prefill and greedily decode. The edited
prefill state remains available to subsequent generated tokens.

In the transport setting, before decoding, we mask attention to the description positions of the prompt that name the factor being intervened on. This prevents prompt instructions from contradicting the intervention, and helps disentangle our representational intervention from token-level effects. For Qwen, which has a recurrent state, we additionally steer the recurrent state of its gated-delta-net layers. Since ablating the key–value rows alone leaves the description alive in that memory, we add to each recurrent state the pooled factor contrast computed from the donors' own recurrent states, overwriting what the ablated description had written there.

\paragraph{Strategy edits.}
A strategy edit, written $\Delta_S[s^\star\leftarrow s]$, uses the same donor-contrast estimator as
function and language edits, averaged over other function--language contexts; $s^\star$ is the target
strategy. Installation starts from a prompt with no strategy specified.

\begin{figure}[H]
    \centering
    \includegraphics[width=0.9\linewidth]{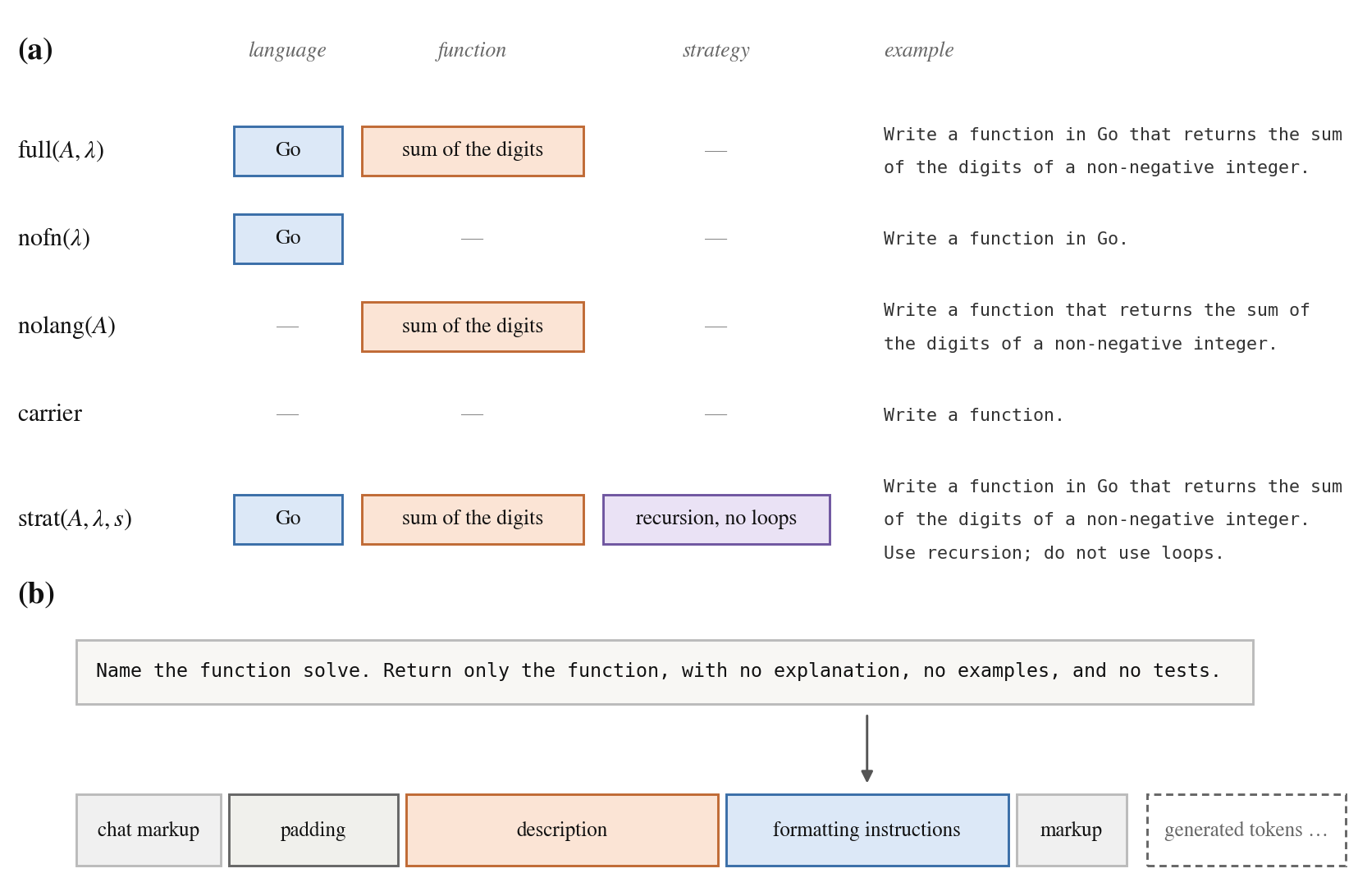}
    \caption{\textbf{(a)}: The different types of prompts and the included factors. \textbf{(b)}: How the prompt is organized.}
    \label{fig:prompts}
\end{figure}

\subsection{Tested Models}
\label{app:llms}

We test a total of 26 LLMs:

\begin{verbatim}
google/gemma-4-31B-it
google/gemma-4-26b-a4b-it
Qwen/Qwen3.6-27B
google/gemma-4-12b-it
Qwen/Qwen3.6-35B-A3B
google/gemma-3-27b-it
Qwen/Qwen3.8-27B
Qwen/Qwen3-30B-A3B-Instruct-2507
Qwen/Qwen3.5-35B-A3B
Qwen/Qwen3-Coder-30B-A3B-Instruct
mistralai/Devstral-2-123B-Instruct-2512
zai-org/GLM-4-32B-0414
google/gemma-3-12b-it
mistralai/Mistral-Small-4-119B-2603
Qwen/Qwen3-32B
Qwen/Qwen3-30B-A3B
mistralai/Mistral-Small-24B-Instruct-2501
LGAI-EXAONE/EXAONE-4.0-32B
mistralai/Mistral-Small-3.2-24B-Instruct-2506
Qwen/Qwen2.5-14B-Instruct
deepseek-ai/DeepSeek-Coder-V2-Lite-Instruct
mistralai/Ministral-3-8B-Instruct-2512
mistralai/Mistral-Nemo-Instruct-2407
google/gemma-2-9b-it
mistralai/Mistral-7B-Instruct-v0.3
allenai/open-instruct-pythia-6.9b-tulu
\end{verbatim}

We compare our breadth-based test and show its correlation with \verb|LiveCodeBench v6| on the 18 out of 26 models that do have public LiveCodeBench scores. 

\begin{figure}[H]
    \centering
    \includegraphics[width=0.9\linewidth]{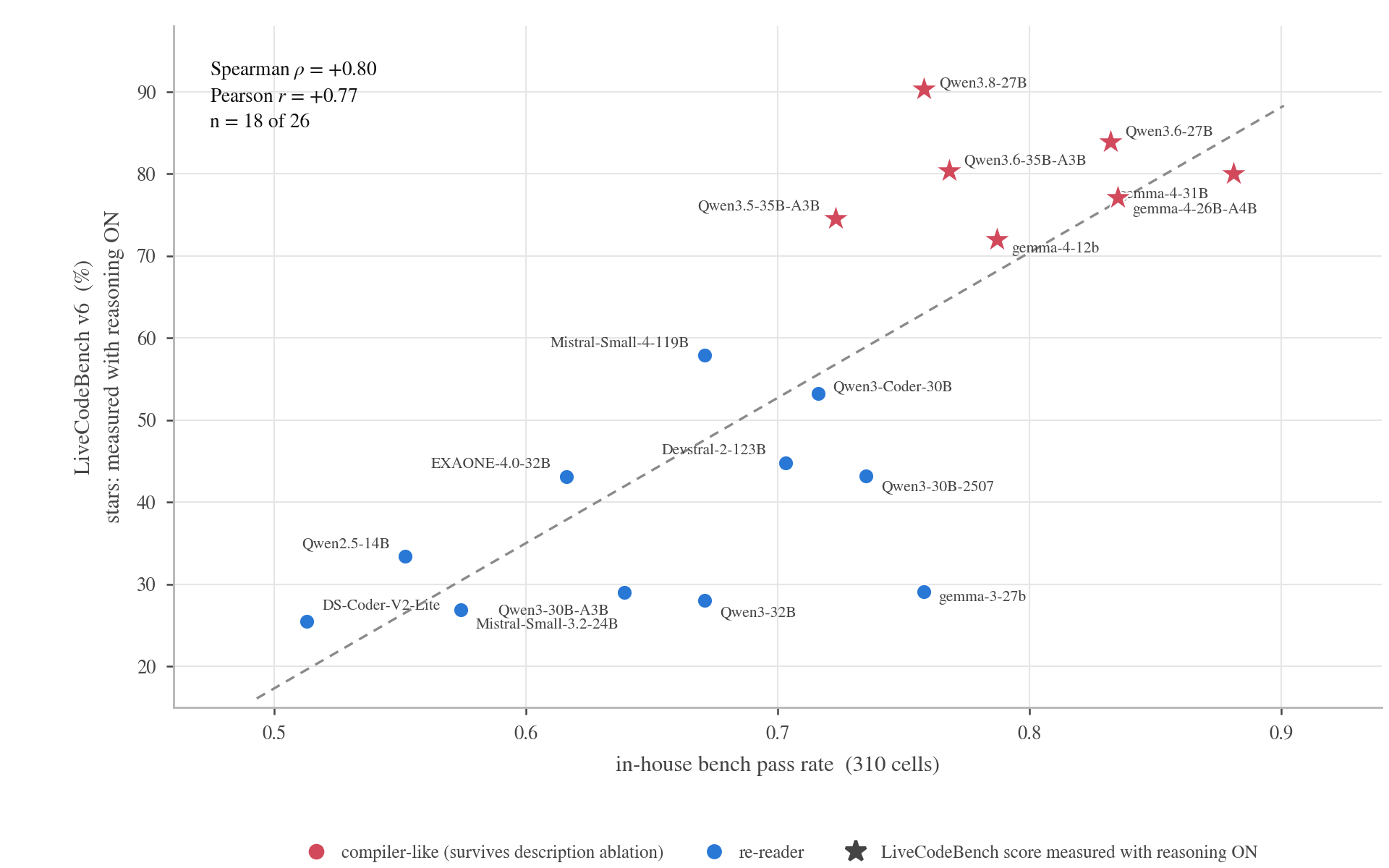}
    \caption{We compare our breadth-based coding benchmark to published scores on LiveCodeBench v6.}
    \label{fig:correlate}
\end{figure}

Figure \ref{fig:correlate} shows the correlation and that the observation that the most capable models tend to be compiler-like still holds. We note that the difference in ordering can possibly be explained by reasoning. \verb|Qwen3.8-27B| being a newer model has likely received more reasoning-based post-training.

\subsection{Programming Languages, Tasks, and Tests}
\label{app:programs}

To test model coding ability, we use 31 tasks where we prompt the model for a solution in every language for 310 total problems. These problems are split into 6 easy, 8 medium, 10 hard, and 7 expert difficulty problems.

\begin{longtable}{@{}l l p{0.58\linewidth}@{}}
\caption{The 31 tasks of the coding-ability bench. Each task is requested in every one of the 10 languages (310 cells per model); a cell passes only if the generated function returns the exact reference output on all six battery inputs. \emph{verlix}, \emph{dromber} and \emph{trelk} ($\dagger$) are invented specifications, so the bank cannot be solved from memory alone.}\label{tab:bench}\\
\toprule
tier & task & specification (as prompted: ``Write a function in $\lambda$ that \ldots'')\\
\midrule
\endfirsthead
\toprule
tier & task & specification (as prompted: ``Write a function in $\lambda$ that \ldots'')\\
\midrule
\endhead
easy & \texttt{digit\_sum} & returns the sum of the digits of a non-negative integer\\
easy & \texttt{count\_divisors} & returns the number of positive divisors of a positive integer\\
easy & \texttt{int\_sqrt} & returns the floor of the square root of a non-negative integer, computed exactly (beware floating point at large inputs)\\
easy & \texttt{largest\_prime\_factor} & returns the largest prime factor of an integer greater than 1\\
easy & \texttt{binary\_gap} & returns the length of the longest run of zeros strictly between two ones in the binary representation of a positive integer\\
easy & \texttt{alt\_digit\_sum} & returns the alternating sum of the digits of a positive integer, starting with a plus sign on the most significant digit\\
\midrule
medium & \texttt{collatz\_steps} & returns the number of Collatz steps (n $\rightarrow$ n/2 if even, else 3n+1) needed to reach 1 from a positive integer\\
medium & \texttt{euler\_phi} & returns Euler's totient of a positive integer\\
medium & \texttt{josephus3} & returns the 1-indexed survivor position of the Josephus problem with n people in a circle, eliminating every 3rd person\\
medium & \texttt{nth\_prime} & returns the n-th prime number, where the 1st prime is 2\\
medium & \texttt{happy\_steps} & returns how many times the sum-of-squared-digits map must be applied to a positive integer before reaching 1 or 4\\
medium & \texttt{partitions} & returns the number of integer partitions of n, modulo $10^9+7$\\
medium & \texttt{binary\_multiple\_digits} & returns the number of decimal digits of the smallest positive multiple of n whose decimal representation contains only the digits 0 and 1\\
medium & \texttt{gray\_weight} & returns the number of ones in the binary reflected Gray code of a non-negative integer\\
\midrule
hard & \texttt{sq\_digit\_sum} & returns the sum of the squares of the digits of a non-negative integer\\
hard & \texttt{even\_digits} & returns how many of the digits of a non-negative integer are even\\
hard & \texttt{digit\_prod} & returns the product of the digits of a non-negative integer\\
hard & \texttt{largest\_proper\_divisor} & returns the largest proper divisor of a positive integer\\
hard & \texttt{derangements} & returns the number of derangements of n elements, modulo $10^9+7$\\
hard & \texttt{distinct\_digit\_count} & returns how many integers from 1 to n have no repeated decimal digit\\
hard & \texttt{fact\_zeros\_base12} & returns the number of trailing zeros of n factorial when written in base 12\\
hard & \texttt{squarefree\_count} & returns how many integers from 1 to n are squarefree (not divisible by any perfect square greater than 1)\\
hard & \texttt{verlix}$^{\dagger}$ & computes the verlix value of n: repeatedly, while n is at least 100, replace n by the sum over its decimal digits of digit$\cdot$(1+i) where i is the digit's position from the right starting at 0; when n drops below 100, return steps$\cdot$1000+n where steps counts the replacements\\
hard & \texttt{dromber}$^{\dagger}$ & computes the dromber value of n: starting with s=0, scan the decimal digits of n from left to right; for each even digit d set s to 2$\cdot$s+d, for each odd digit d set s to s+d$\cdot$d; return s modulo $10^5$\\
\midrule
expert & \texttt{mountain\_count} & returns how many integers from 1 to n have decimal digits that strictly increase and then strictly decrease, with both parts non-empty\\
expert & \texttt{fib\_reps} & returns the number of ways to write n as a sum of distinct Fibonacci numbers from the sequence 1, 2, 3, 5, 8, ... (each used at most once)\\
expert & \texttt{largest\_palindrome\_leq} & returns the largest integer not exceeding n whose decimal representation is a palindrome\\
expert & \texttt{oddfact\_zeros\_base15} & returns the number of trailing zeros, in base 15, of the product 1$\cdot$3$\cdot$5$\cdot$...$\cdot$(2n-1) of the first n odd numbers\\
expert & \texttt{trelk}$^{\dagger}$ & computes the trelk value of n: starting with s=0 in mode A, scan the decimal digits left to right; a zero digit switches the mode (A to B or B to A) and is otherwise ignored; in mode A a digit d sets s to s+d$\cdot$d, in mode B it sets s to 2$\cdot$s+d; return s modulo $10^6$\\
expert & \texttt{josephus\_sqrt} & returns the 1-indexed survivor of the Josephus problem with n people where every k-th person is eliminated and k is the floor of the square root of n\\
expert & \texttt{prime\_digitprime\_next} & returns the smallest prime strictly greater than 2n whose digit sum is also prime\\
\bottomrule
\end{longtable}

We use 10 languages in total (8 imperative, 2 \textit{functional}). Those are:

\begin{center}
\begin{tabular}[H]{llll}
Python & JavaScript & C     & C++ \\
Ruby   & Java       & Go    & Rust \\
\textit{OCaml}  & \textit{Haskell}    &       &
\end{tabular}
\end{center}

We choose a smaller subset for intervention experiments:

\begin{center}
\begin{tabular}[H]{ll}
\verb|digit_sum| & \verb|digit_prod|\\
\verb|even_digits|   & \verb|sq_digit_sum| \\
\end{tabular}
\end{center}

For every realized program, we run 6 unit tests and measure success only if all tests pass.

\subsection{In-context learning}
\label{app:icl}

This section expands on Figure~\ref{fig:icl-sft}a; Table~\ref{tab:install_per_algo} gives the results per algorithm and model.

In this setting, a prompt shows one program in language $\lambda$ (``Here is a function:'' followed by the program) and the shared suffix ``Write the function above. It takes a whole number and returns an integer. Use the same algorithm. Name the function solve. Return only the code''.

As in Section~\ref{sec:shared-interventions}, we mask attention to the example after the suffix, so that only the 39 suffix positions can carry the function. A pair \emph{compiles} if the decoded program still passes under this intervention.

 We then harvest the suffix state of a donor and write it into the recipient's suffix cache with $\alpha = 1$, append tokens requesting language $\lambda$, and decode. We only consider donors that successfully compiled. \emph{Single donor} transplants one donor's state; \emph{mean over donors} uses the mean state of the admissible donors. Controls are the recipient alone and the recipient with matched-norm Gaussian noise added to the suffix cache.

\begin{table}[H]
\caption{In-context transplant per algorithm. \emph{compiles}: donor--target pairs (of 90) whose suffix state still yields the function when attention to the example is masked (RWKV-7: unmasked pairs). \emph{no example}: the recipient alone; \emph{mean over donors}: the mean state of the admissible donors, one cell per target language; \emph{single donor}: one donor's state per (admissible donor, target) cell. A cell passes only if all six unit tests pass in the requested language. The random-edit control passes 0 cells.}
\label{tab:install_per_algo}
\begin{center}
\small\setlength{\tabcolsep}{5pt}
\begin{tabular}{l l l r r r r}
\toprule
\multicolumn{1}{c}{\bf model} & function & kind & compiles & no example & mean over donors & single donor \\
\midrule
Gemma-4-31B & \texttt{thulep} & regular & 87/90 & 0/10 & 10/10 & 87/87 \\
Gemma-4-31B & \texttt{vantril} & regular & 81/90 & 0/10 & 10/10 & 80/81 \\
Gemma-4-31B & \texttt{kanex} & nonce & 50/90 & 0/10 & 0/10 & 42/50 \\
Gemma-4-31B & \texttt{faromin} & nonce & 64/90 & 0/10 & 1/10 & 54/64 \\
\midrule
Gemma-4-26B-A4B & \texttt{thulep} & regular & 80/90 & 0/10 & 10/10 & 80/80 \\
Gemma-4-26B-A4B & \texttt{vantril} & regular & 77/90 & 0/10 & 8/10 & 72/77 \\
Gemma-4-26B-A4B & \texttt{kanex} & nonce & 71/90 & 0/10 & 9/10 & 69/71 \\
Gemma-4-26B-A4B & \texttt{faromin} & nonce & 32/90 & 0/9 & 4/9 & 27/32 \\
\midrule
Qwen3.6-27B & \texttt{thulep} & regular & 87/90 & 0/10 & 10/10 & 86/87 \\
Qwen3.6-27B & \texttt{vantril} & regular & 50/90 & 0/10 & 8/10 & 48/50 \\
Qwen3.6-27B & \texttt{kanex} & nonce & 27/90 & 0/10 & 10/10 & 26/27 \\
Qwen3.6-27B & \texttt{faromin} & nonce & 13/90 & 0/8 & 7/8 & 13/13 \\
\midrule
RWKV-7 13B & \texttt{thulep} & regular & 80/90 & 0/10 & 10/10 & 80/80 \\
RWKV-7 13B & \texttt{vantril} & regular & 75/90 & 0/10 & 10/10 & 75/75 \\
RWKV-7 13B & \texttt{kanex} & nonce & 81/90 & 0/10 & 9/10 & 81/81 \\
RWKV-7 13B & \texttt{faromin} & nonce & 71/90 & 0/10 & 8/10 & 71/71 \\
\bottomrule
\end{tabular}
\end{center}
\end{table}

We see in Table \ref{tab:install_per_algo} that Single-donor transplants pass in 93--100\% of addressable cells on every model and the controls in none. The mean over donors is nearly exact for the regular algorithms (76/80 targets) but not for the nonces: on Gemma-4-31B the mean kanex state yields \texttt{digit\_sum} in all ten targets and the mean faromin state passes 1/10, while their single donors pass 42/50 and 54/64. The mean keeps the shared computation and drops the nonce's extra step.

\subsection{Supervised Fine-Tuning Parameters and Data}
\label{app:sft}

For the full SFT experiments we use 4 algorithms specified in table \ref{tab:digit-properties}.

\begin{table}[H]
\centering
\caption{Digit-based properties of a non-negative integer $n$.}
\label{tab:digit-properties}
\begin{tabular}{lp{0.72\linewidth}}
\hline
\textbf{Name} & \textbf{Definition for a non-negative integer $n$} \\
\hline
\texttt{vantril} & Number of distinct decimal digits appearing in $n$. \\
\texttt{thulep}  & Largest decimal digit of $n$ minus the smallest decimal digit of $n$. \\
\texttt{kanex}   & Sum of $d+1$ over all decimal digits $d$ of $n$ (the digit sum plus the number of digits). \\
\texttt{faromin} & Sum of $d+1$ over the distinct decimal digits $d$ of $n$, each digit value counted once. \\
\hline
\end{tabular}
\end{table}

We use 3 paraphrases and 3 implementations, all in Python, for a total of 9 training examples. The full paraphrases and implementations can be seen at \url{https://github.com/Astera-org/code-abstractions}.

\begin{table}[H]
\centering
\caption{Fine-tuning hyperparameters used across experiments.}
\label{tab:hyperparameters}
\begin{tabular}{ll}
\toprule
\textbf{Hyperparameter} & \textbf{Value} \\
\midrule
Optimizer & AdamW \\
LR schedule & Cosine decay \\
Warmup & 3\% \\
Gradient clipping & 1.0 \\
Precision & bf16 \\
Sharding & DeepSpeed ZeRO-3 \\
Sequence length & 448 \\
Seed & 0 \\
Batching & One step per epoch over the 9 examples \\
\bottomrule
\end{tabular}
\end{table}

\begin{table}[H]
\centering
\caption{Fine-tuning dosage sweep.}
\label{tab:dosage}
\begin{tabular}{ccc}
\toprule
\textbf{Dose} & \textbf{Learning rate} & \textbf{Epochs} \\
\midrule
$1\times$  & $5\times10^{-7}$ & 36 \\
$4\times$  & $2\times10^{-6}$ & 36\\
$16\times$ & $8\times10^{-6}$ & 36 \\
\bottomrule
\end{tabular}
\end{table}

We use the parameters specified in table \ref{tab:hyperparameters} and additionally sweep 3 different learning rates for each model. We use the next-token prediction objective and calculate loss only on the assistant output positions.

\begin{figure}[H]
    \centering
    \includegraphics[width=0.9\linewidth]{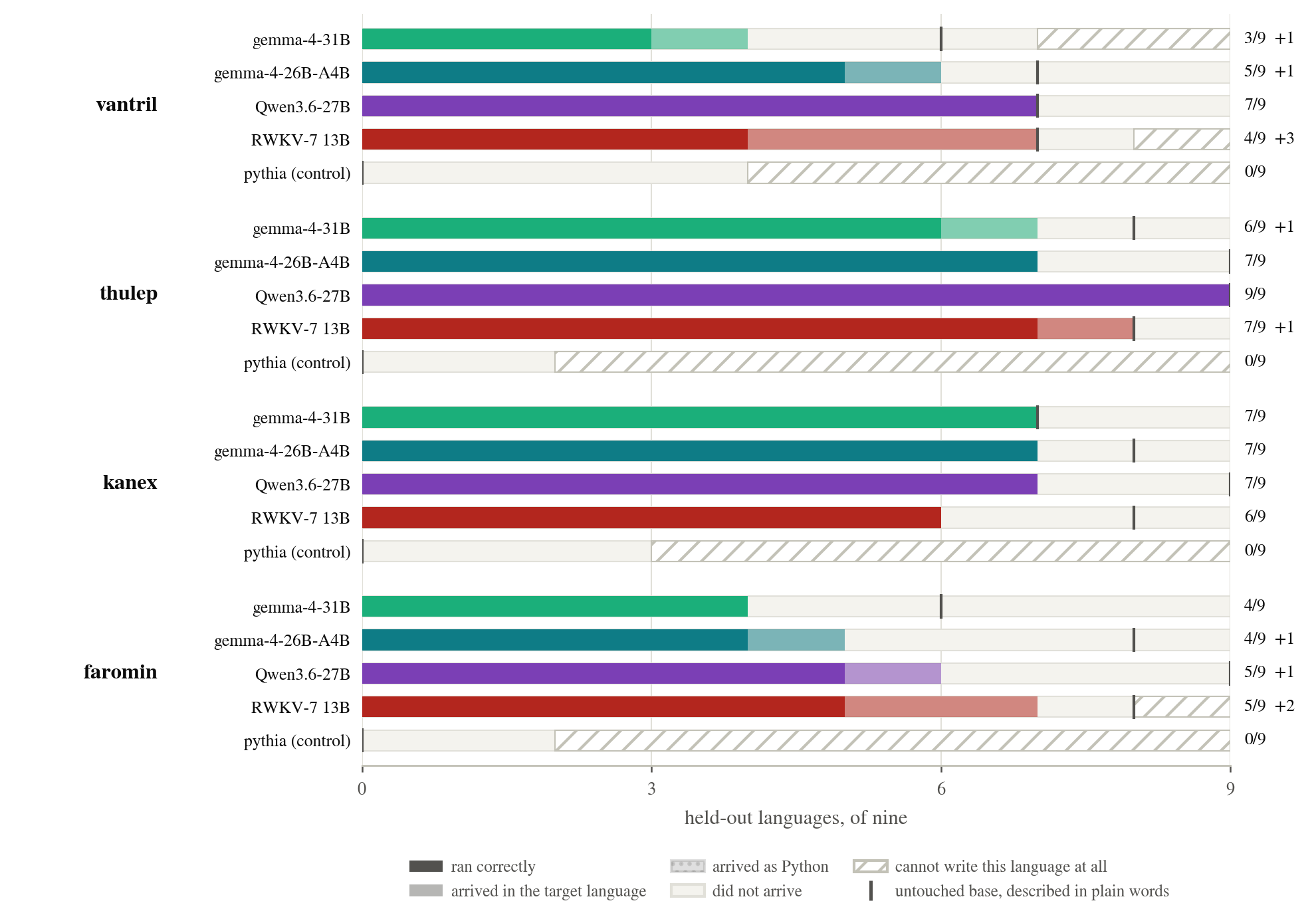}
    \caption{The full generalization results across all the invented algorithms.}
    \label{fig:sft-extend}
\end{figure}

We then evaluate each model's generalization to the 9 unseen languages. In general, we see in figure \ref{fig:sft-extend} that the capable models have a transfer capability: each algorithm is consistently implemented in more than a single language. In contrast, our control model shows no sign of generalization. It learns the algorithm in Python, but even within the other languages that it does know, it does not implement the newly taught algorithm. Note that generalization in general is a property of how well-optimized the training was. Given that we are additionally fine-tuning an already post-trained model with very little data, we expect that generalization can be better if the models are more carefully optimized. Our aim here is to show that a model can generalize from very little data; not to show how to train models in low-data regimes.

\subsection{Compute Resources}

We used a GPU server with 8x H100 Nvidia GPUs with ~82 GB of VRAM each. With DeepSpeed ZeRO-3, we were able to fine-tune all of the models without having to resort to Parameter-Efficient Fine-Tuning strategies.

Experiments under different language/function/strategy settings were run serially. Specifically, no batch would ever contain multiple different programming languages or functions. Batching can cause differences in output probabilities due to ordering, which may cause differences when using greedy decoding.

\subsection{Pretraining vs. Instruct-Tuning}
\label{app:pretrain}
We choose instruct-tuned models for 2 reasons: 
\begin{enumerate}
    \item Most open-source models now ship as instruct-tuned models by default
    \item They are easier to prompt
\end{enumerate}

However, this introduces a question of whether the abstraction we measure at the prompt suffix is a result of instruct-tuning or if it exists from pre-training alone. Here we compare the results of ablation on pretrained models (where available) vs. their instruct-tuned counterparts.

\begin{figure}[H]
    \centering
    \includegraphics[width=0.9\linewidth]{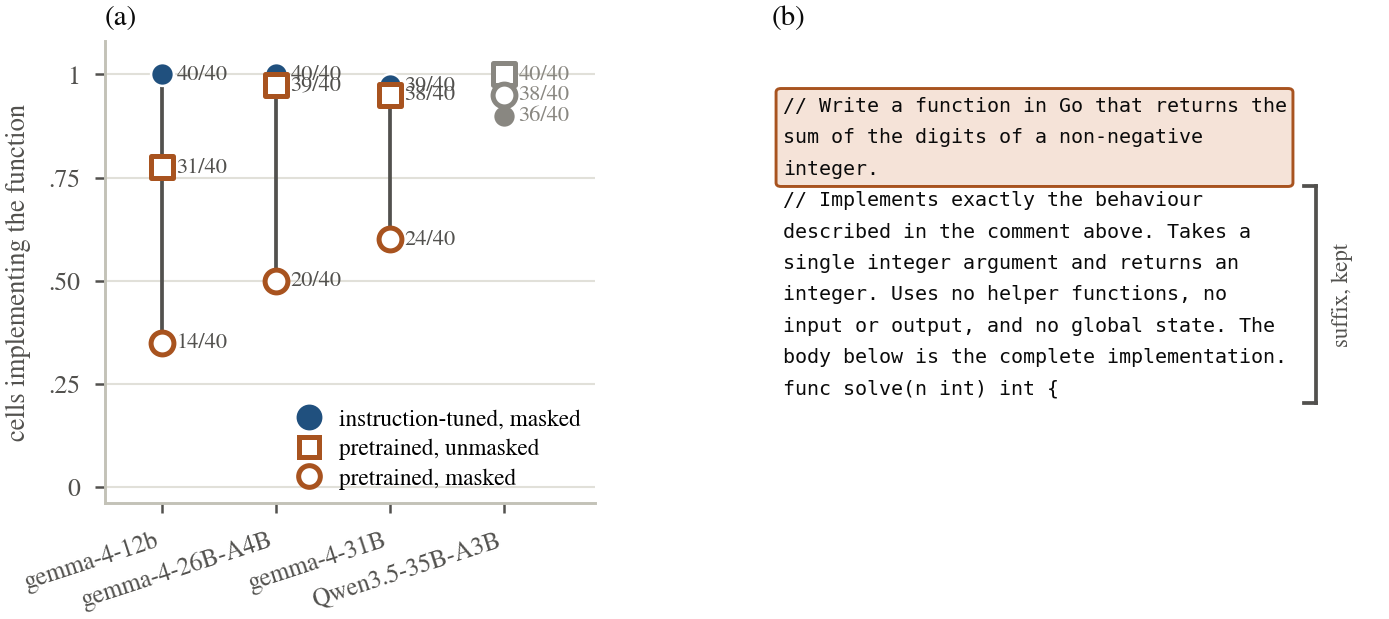}
    \caption{\textbf{(a)}: Pretrained models are substantially worse than their instruct-tuned counterparts at generating correct code with a masked out description. \textbf{(b)}: The prompt we use for the base-models to trigger their completion.}
    \label{fig:pretrain}
\end{figure}

Figure \ref{fig:pretrain} shows that while pretrained models do show signs of being able to solve some tasks without direct access to the description, the proportion is less than that of the instruct-tuned ones. We note that a possible confounder is how pretrained models respond, and the difference in how we must prompt them as a result. Regardless, it does seem like post-training induces more pressure for the model to create a persistent abstraction.

Note that the most unaffected model, Qwen, also has a recurrent state due to its linear attention. This naturally forces the model to retain some abstract state of its instructions in a way that regular attention models do not have to. Importantly, we cannot fully disentangle this effect from the additional code training the model likely sees during post-training, however we think it is suggestive that the performance drop when masking the base model is substantially higher.

\subsection{Addressability}
\label{app:addressability}
One point we highlight multiple times in the main paper is that our interventions are largely based on finding the correct way to address a given model.

To illustrate this, we replicate the main figure experiments from section \ref{sec:shared-interventions} on \verb|Mistral-Small-3.2| in 4 different ways:
\begin{enumerate}
    \item At the suffix
    \item At the suffix with attention to the description masked
    \item At generation time with a static vector
    \item At generation time steering the whole trajectory
\end{enumerate}

Both generation-time interventions add to the residual stream at every layer
of the token being written, at the launch token and at every generated token,
and use the same leave-one-out donor pools and the same controls (norm-matched
random direction, wrong value, mean over values, random labelling) as the
suffix interventions. They differ only in how the direction is obtained.

\paragraph{\textbf{Static}:} For each donor pair $(x, y)$ in the pool, the program
    the model writes for $x$ is teacher-forced through both prompts; the
    residual difference $h_\ell(x, \cdot) - h_\ell(y, \cdot)$ is averaged over
    the program's positions, then over pairs (and summed over factors for
    composition). The resulting per-layer vector is fixed for the cell and
    added with coefficient $\alpha = 0.25$, the best of a $0.0625$--$1.0$
    sweep. This is the standard steering-vector construction, read at the
    continuation positions.

\paragraph{\textbf{Dynamic}:} The direction is recomputed at every generated
    token: each donor prompt in the pool is teacher-forced on the recipient's
    own output so far, $y_{<t}$, and the added vector is the pooled difference
    $\delta_{\ell,t} = \mathrm{mean}_{(x,y)}\,[\,h_\ell(x, y_{<t}) - h_\ell(y, y_{<t})\,]$,
    with $\alpha = 0.0625$ ($0.125$ for the strategy factor). Random controls
    are norm-matched to $\delta_{\ell,t}$ at each step. This is the same
    contrast the suffix intervention writes once into the K/V rows, delivered
    at the position where this model reads it.

\begin{figure}[H]
    \centering
    \includegraphics[width=0.9\linewidth]{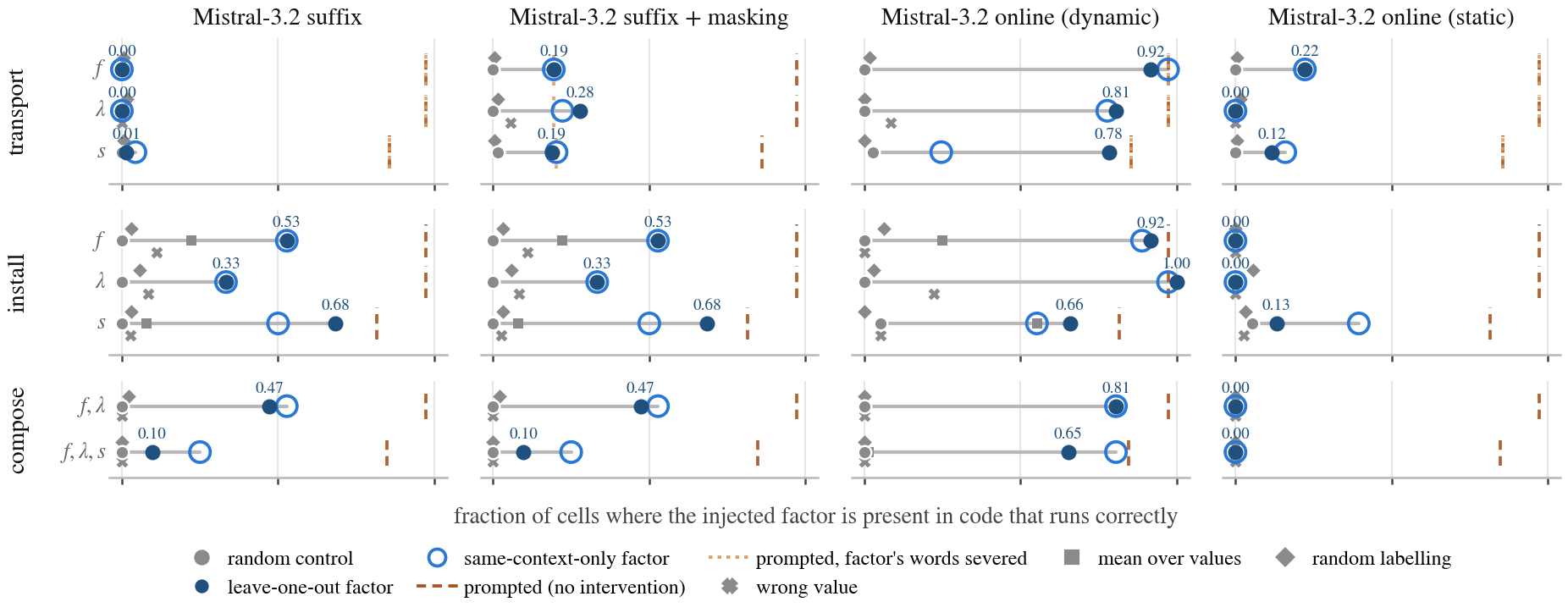}
    \caption{On Mistral-Small-3.2, the same intervention site that worked well on the main models largely fails.}
    \label{fig:mistral}
\end{figure}

Figure \ref{fig:mistral} shows the result of all 4 interventions. Note especially that the suffix-interventions that were largely successful in the main paper models are ineffective here; especially under transport. With no masking, the model sees the factor tokens in the prompt and the suffix steering is not enough to overwrite them. With masking, the intervention success increases but the overall coding performance plummets. The takeaway is that we cannot rule out abstractions in models that do not store reusable representations at their suffix; we simply must address them differently. They may sit somewhere between the suffix, the description tokens, already-generated tokens, or some mixture of all of the above.

\subsection{What is the Compiled State?}
\label{app:compiler}
We choose \verb|Gemma-4-31B-it| for this investigation and provide additional clarification on what the ``compiled'' state at the prompt suffix is.

To investigate this, we ask the model to implement a function that sums the digits of an integer in Python. Additionally, we maintain a pool of 11 extra instructions that we add to the descriptions. These range from specifying the strategy (iteration vs. recursion) to requesting certain intermediary variable names or comments. We then consider 12 different increasing suffix lengths and for each, sample 10 trials of $n$ different instructions from 1 to 10. 

\begin{figure}[H]
    \centering
    \includegraphics[width=0.9\linewidth]{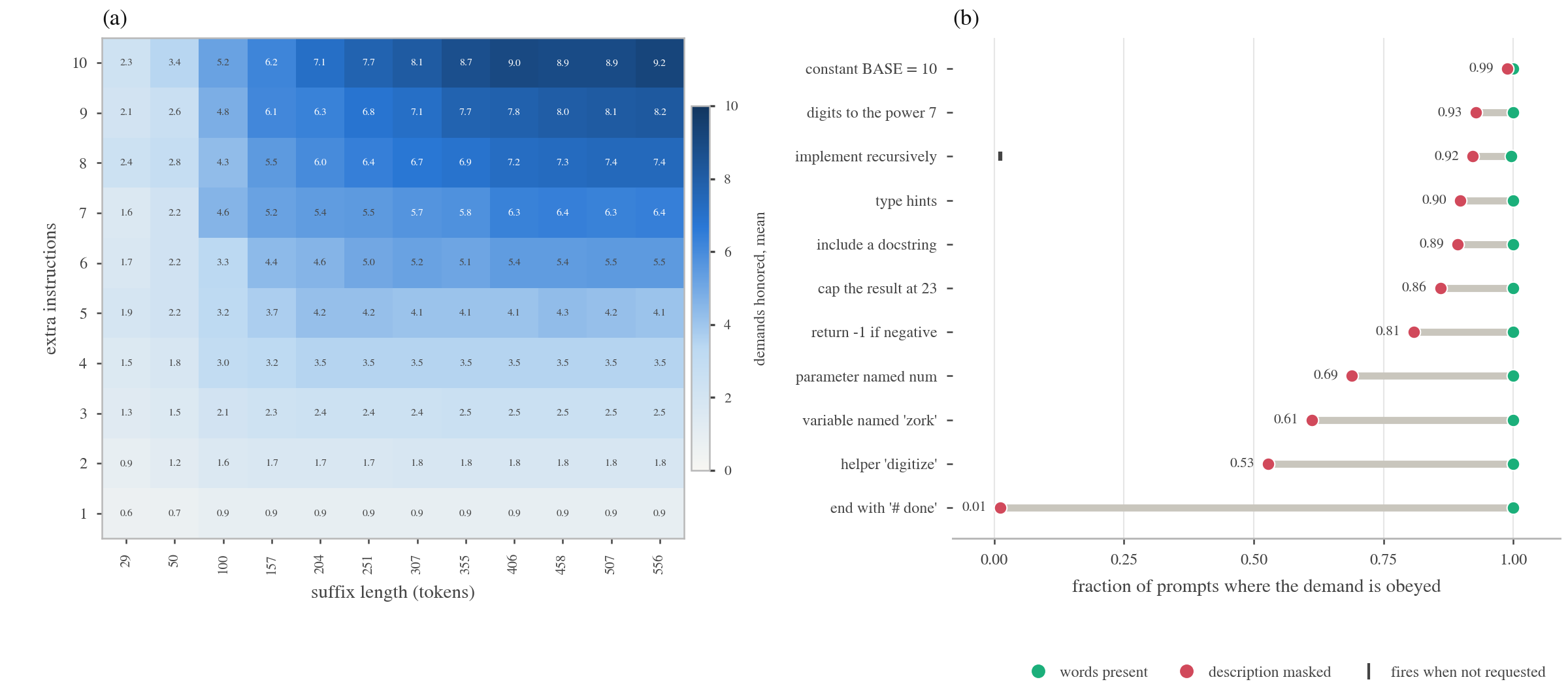}
    \caption{\textbf{(a)}: As more instructions are provided, a longer suffix length is needed for the model to store the program state. \textbf{(b)}: Not all instructions are equal. Instructions less relevant for computation are more often early re-read as opposed to being part of the computation state. With an unablated description, all instructions are honored.}
    \label{fig:difficulty}
\end{figure}

Figure \ref{fig:difficulty}a shows that a larger number of extra instructions demands a longer suffix length to compile into a valid program abstraction. Additionally, we see in figure \ref{fig:difficulty}b that certain instructions generally compile less than others, and thus are more generally re-read as required rather than carried into the program state. For highly complex programs, it is clear that the model relies on a combination of its abstract program state and eagerly re-reading instructions.

\subsubsection{Links to Reasoning}
The construction in Appendix~\ref{app:toy-process} suggests a possible mechanism by which reasoning
training could strengthen compiler-like internal representations.

The toy separates a fixed task condition $\theta=(f,\lambda)$ from its evolving machine states.
A blank $\square$ leaves the function and watcher states unchanged, and its likelihood is independent
of $\theta$. For every supported length-$B$ extension $h\square^B$ of context history $h$, the exact task belief
$\eta_{\mathrm{task}}=\eta_L\otimes\eta_F$ is therefore preserved:
\[
    \eta_{\mathrm{task}}(h\square^B)=\eta_{\mathrm{task}}(h).
\]
This statement concerns Bayesian beliefs about the source, not equality of a trained model's activations.
It motivates a model retaining task information across a region that supplies no further task evidence.

We hypothesize that reasoning tokens may create analogous pressure. They are not literally blank
emissions: reasoning can perform substantial computation and introduce intermediate variables. However,
many steps preserve the identity of the task being solved. A persistent activation state could therefore
decompose schematically into a task contribution and an evolving workspace,
\[
    Z_t\approx Z_{\mathrm{task}}\oplus Z_{\mathrm{work},t},
\]
with the task contribution retained while the workspace changes:
\[
    (Z_{\mathrm{task}},Z_{\mathrm{work},t})
    \longrightarrow
    (Z_{\mathrm{task}},Z_{\mathrm{work},t+1}).
\]
The hypothetical neural contribution $Z_{\mathrm{task}}$ need not be an exact encoding of
$\eta_{\mathrm{task}}$.

This suggests a possible origin for the compiler-like mechanism observed
above. Long reasoning trajectories repeatedly require a model to preserve
task constraints while performing many intervening computations. Reasoning
training may therefore reward a division of labor in which the specification
is ``compiled'' into a persistent latent state, while a separate state evolves
to carry out the computation:
\[
    \text{specification}
    \longrightarrow
    \underbrace{Z_{\mathrm{task}}}_{\text{persistent compiled state}}
    \longrightarrow
    \underbrace{Z_{\mathrm{work},0}\to\cdots\to Z_{\mathrm{work},T}}
                _{\text{reasoning computation}}
    \longrightarrow
    \text{realization}.
\]

We emphasize that this is a hypothesis rather than a conclusion of the
present experiments. The experiments here study models without reasoning;
the possibility that factored representations persist or evolve throughout a
reasoning trace is left open. This makes reasoning-trained models a
particularly interesting setting in which to test whether extended
computation strengthens the persistence, factorization, or causal
transportability of compiled task state. 

A direct prediction is that, conditional on the same task specification,
reasoning-trained models should retain recoverable information about factors
such as function, language, and strategy across longer stretches of intervening computation,
including positions whose local token content contains little direct evidence
about those factors. Note that this applies even to Transformer models with attention. As context length grows, it becomes harder for a few early tokens (for instance, initial instructions) to ``win'' the softmax and transport information. We do not claim that this is the only mechanism by which reasoning works, simply that it is one by which a compiled state may be computationally encouraged.

\subsection{Strategy Steering}
\label{app:strategy}
We expand on the results in figure \ref{fig:abstraction} and for each model, extend this result.

For each language $\times$ language pair, we ask what a strategy donor from one language realizes in the other. Specifically we investigate how successful steering is from single donors and when it is, whether the steered realization matches prompting the donor, recipient, or something else, for cases where the recipient and donor disagree on the computation.

\begin{figure}[H]
    \centering
    \includegraphics[width=0.98\linewidth]{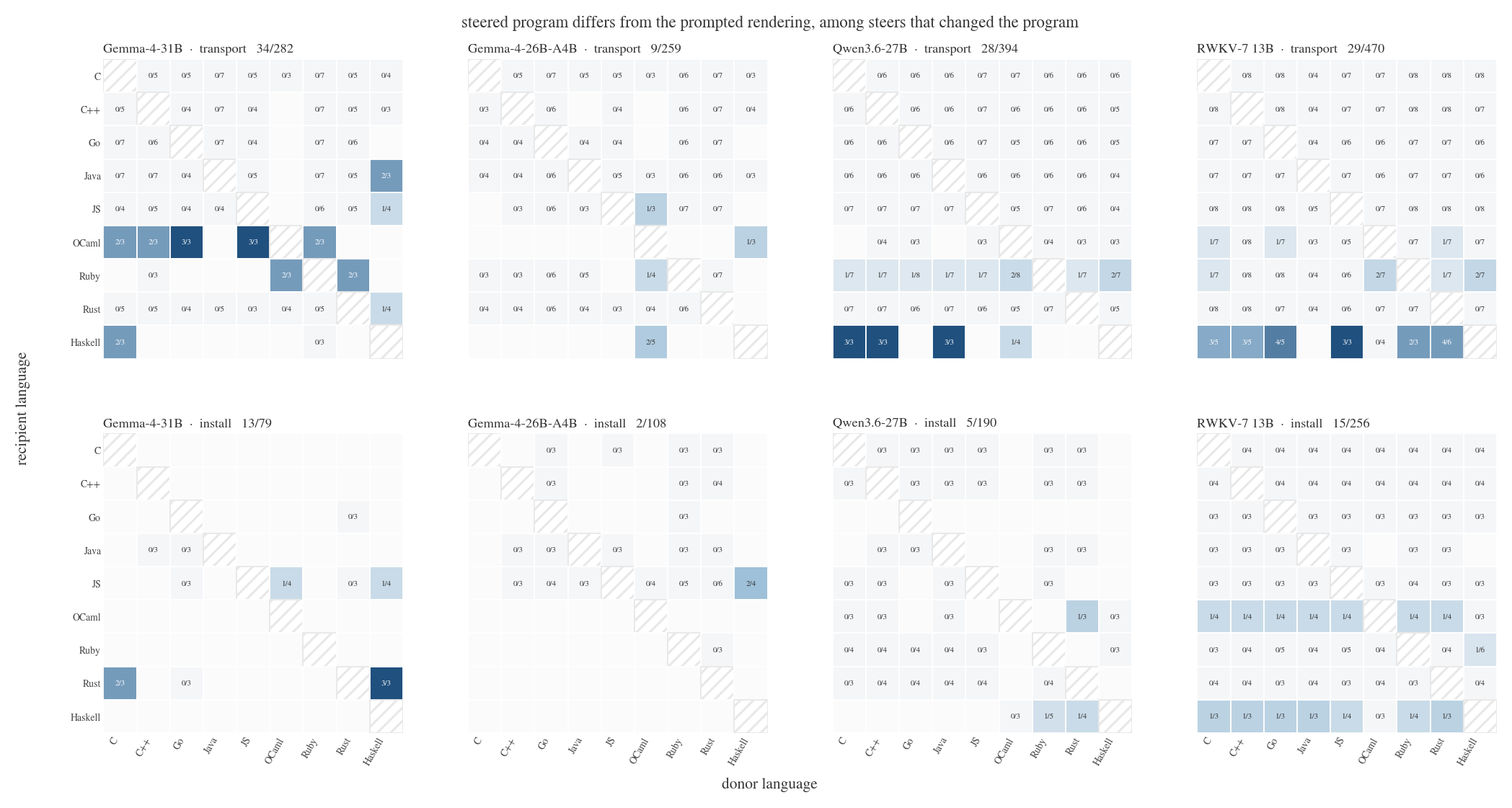}
    \caption{For each model pair and intervention, when strategy steering succeeds from a single donor, how is the resulting program expressed? Blank squares imply strategy steering did not work.}
    \label{fig:strategy}
\end{figure}

\begin{figure}[H]
    \centering
    \includegraphics[width=0.9\linewidth]{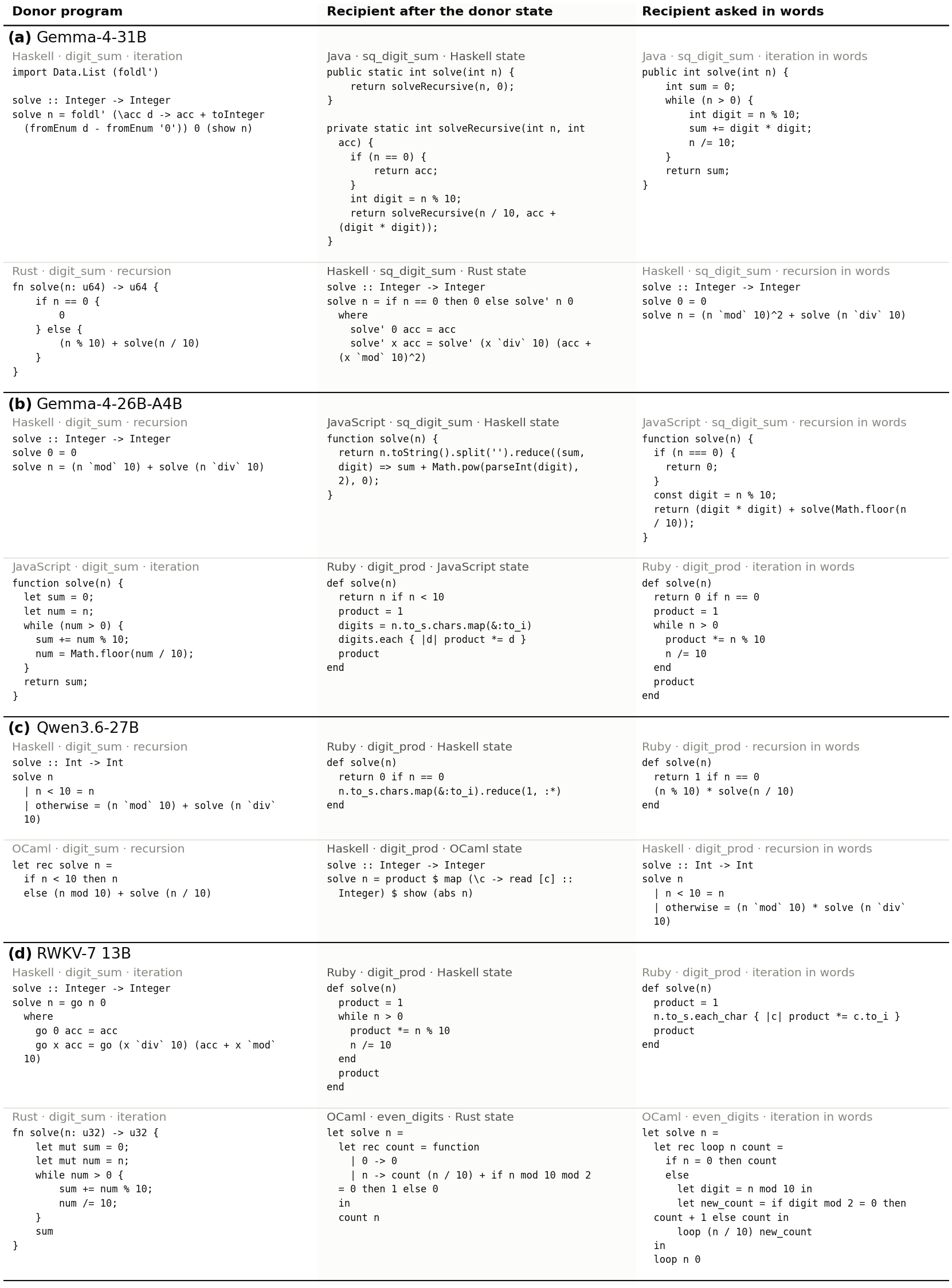}
    \caption{For each model, concrete examples where the steered recipient expresses a program differently from when it is prompted for the same strategy.}
    \label{fig:strategy-examples}
\end{figure}

Figure \ref{fig:strategy} shows that successful strategy-steering from a single donor heavily depends on both the target and recipient language. Additionally, the steering sometimes causes realizations in forms that are different from how the recipient would normally write a given program if prompted to do it.  

These results suggest that the interventions are not merely causing rewrites of the model prompts. Rather, when the models see a description containing a function and a language, the compiled state can also contain a language-specific intended realization of that computation. When we steer, this realization may sometimes carry over---more frequently when the languages differ substantially, as is the case with imperative vs. functional languages.

\subsection{Local Geometry}
\label{app:geometry}
\begin{figure}[H]
    \centering
    \includegraphics[width=0.9\linewidth]{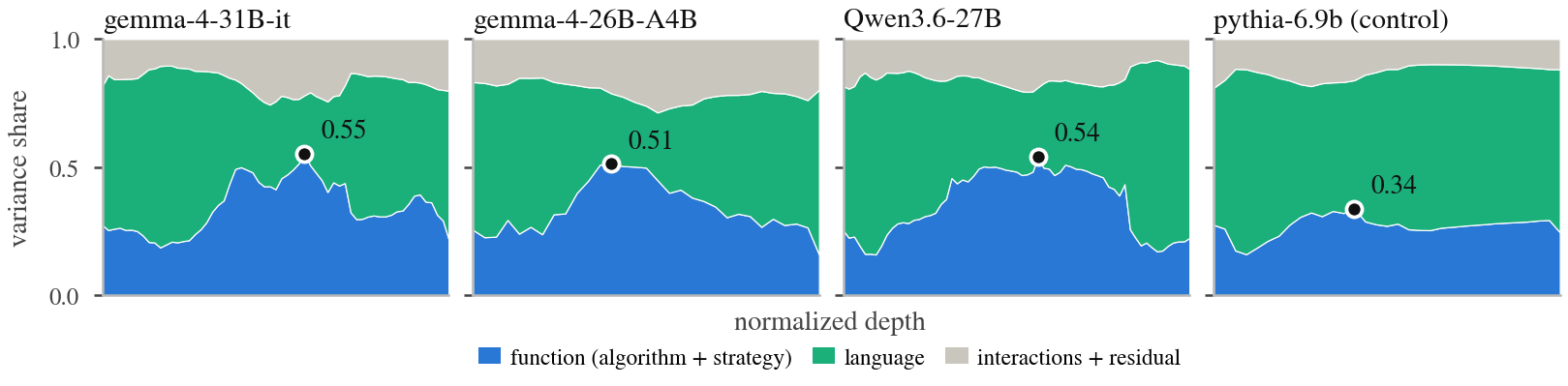}
    \caption{In the middle layers of the tested models, the function being implemented accounts for a larger share of the variance than the language it is implemented in.}
    \label{fig:representation-depth}
\end{figure}

There are two important consequences of the way we specify equation \ref{eq:code-factorization}. First, no single layer has to realize the entire factorization. Different factors may be constructed and used at different depths. Second, if factors are reused, then at some point in the network variation in e.g. function identity should be attributable primarily to the function's computation rather than its lexical properties. We let each model solve a subset of 20 programs from our corpus in every language. We then stack and decompose those activations at every layer. Figure \ref{fig:representation-depth} shows the aforementioned consequences clearly; towards the middle of the model, computation-related main effects become a major source of variance, whereas language matters more before and after. Our control model shows signs of the same behavior, but its variance is consistently dominated by variation in language. 
\subsubsection{Interventions}

The interventions in the main paper show that function changes can be reused across contexts, but do not require function state to have globally linear coordinates. We now design interventions to probe the local geometry instead. First, let $Z(f)$ denote the measured state for function $f$ with the remaining prompt context fixed.

We then ask whether a displacement estimated from one pair of functions can act as the same operation elsewhere. From a single donor pair we construct
\begin{equation}
\Delta_{\mathrm{sq}}
=
Z(\mathrm{sq\cdot sum})-Z(\mathrm{sum}),
\end{equation}
and add it to the states for product and maximum. Neither $\mathrm{sq\cdot prod}$ nor $\mathrm{sq\cdot max}$ is used to construct the edit. We vary the dose $\alpha$, decode in four programming languages, and compare against a norm-matched random perturbation. We run these experiments on \verb|Gemma-4-31B|.

\begin{figure}[H]
\centering
\includegraphics[width=0.98\linewidth]{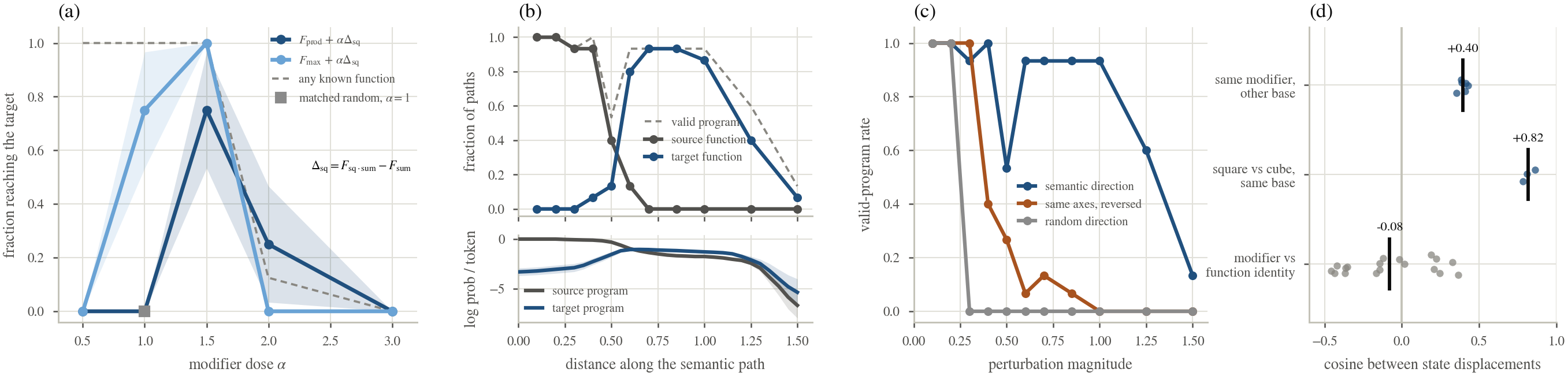}
\caption{
\textbf{Function state has locally reusable but anisotropic causal geometry.}
\textbf{(a)} A square modifier estimated only from sum induces squared-product and squared-max at intermediate dose; a norm-matched random edit does not.
\textbf{(b)} Moving along the displacement between two naturally elicited function states produces a sharp transition from source to target behavior, with validity falling near the transition and under over-extrapolation.
\textbf{(c)} Valid computation is preserved much farther along the semantic direction than along its reversal or a matched random direction.
\textbf{(d)} The same modifier across different bases is aligned, related modifiers on the same base are more strongly aligned, while modifier directions are approximately orthogonal to generic function differences. $F_{\mathrm{sum}},F_{\mathrm{prod}},F_{\mathrm{max}}$ denote the corresponding measured $Z$ states.}
\label{fig:anisotropic}
\end{figure}

Figure~\ref{fig:anisotropic}a shows that the same $\Delta_{\mathrm{sq}}$ can produce both unseen modified functions, but only over a limited range of doses. This looks more like a reusable local operation than a global translation in function space.

We next move directly between two naturally occurring function states. For source $f$ and target $f^\star$, we traverse
\begin{equation}
\widetilde Z(\alpha)
=
Z(f)+\alpha\left[Z(f^\star)-Z(f)\right].
\label{eq:interp}
\end{equation}
Here $\alpha=0$ recovers the source and $\alpha=1$ the same-context target state. Figure~\ref{fig:anisotropic}b shows a sharp transition between the two computations: near the boundary, validity drops, but as soon as it is traversed, valid programs are once again constructed.

We then compare the same semantic direction with its reversal and a norm-matched random direction orthogonal to it; directions that have \textit{not} been shown to result in valid computation whatsoever. Figure~\ref{fig:anisotropic}c shows that this really matters. Traversing along the semantically-valid direction produces correct programs for a much higher degree of perturbation than the transverse and reverse. The picture is mixed: locally, computational directions can be used (and sometimes combined), but the effect is strongly dose-dependent. Arbitrary directions generally do not lead to valid programs.

\subsubsection{Generalization of computational modifiers}

We next ask whether this kind of modifier reuse extends beyond one donor pair and one model. We test the three selected language models and \verb|RWKV7-13.3B| using several transformations, base computations, languages, and specification wordings. For RWKV, we intervene on its recurrent state rather than a KV-cache.

Let $M[f]$ be a transformation of computation $f$, and write $Z(f,\lambda,w)$ for the measured persistent state for function $f$, language $\lambda$, and wording $w$. For each recipient $c_\star=(f_\star,\lambda_\star,w_\star)$, we estimate a modifier only from donors with a different base function, language, and wording:
\begin{align}
\widehat{\Delta}{M}^{(-c\star)}
&=
\frac{1}{|\mathcal D_\star|}
\sum_{(f,\lambda,w)\in\mathcal D_\star}
\left[Z(M[f],\lambda,w)-Z(f,\lambda,w)\right],\
\widetilde Z_\star
&=
Z(f_\star,\lambda_\star,w_\star)
+\alpha\widehat{\Delta}{M}^{(-c\star)}.
\label{eq:modifier}
\end{align}
Importantly, we use a leave-one-out design: the modifier does not contain the recipient function in modified form or examples from the recipient language.

We use sum, product, maximum, minimum, and alternating sum as digit aggregations and pick four modifiers: squaring the aggregate, aggregating squared digits, incrementing it, or doubling it. We use Python and JavaScript to select settings which we re-use for other languages. In total, there are 160 evaluation cells per model. Similarly to the main paper, success requires passing unit tests.

\begin{figure}[H]
\centering
\includegraphics[width=0.9\linewidth]{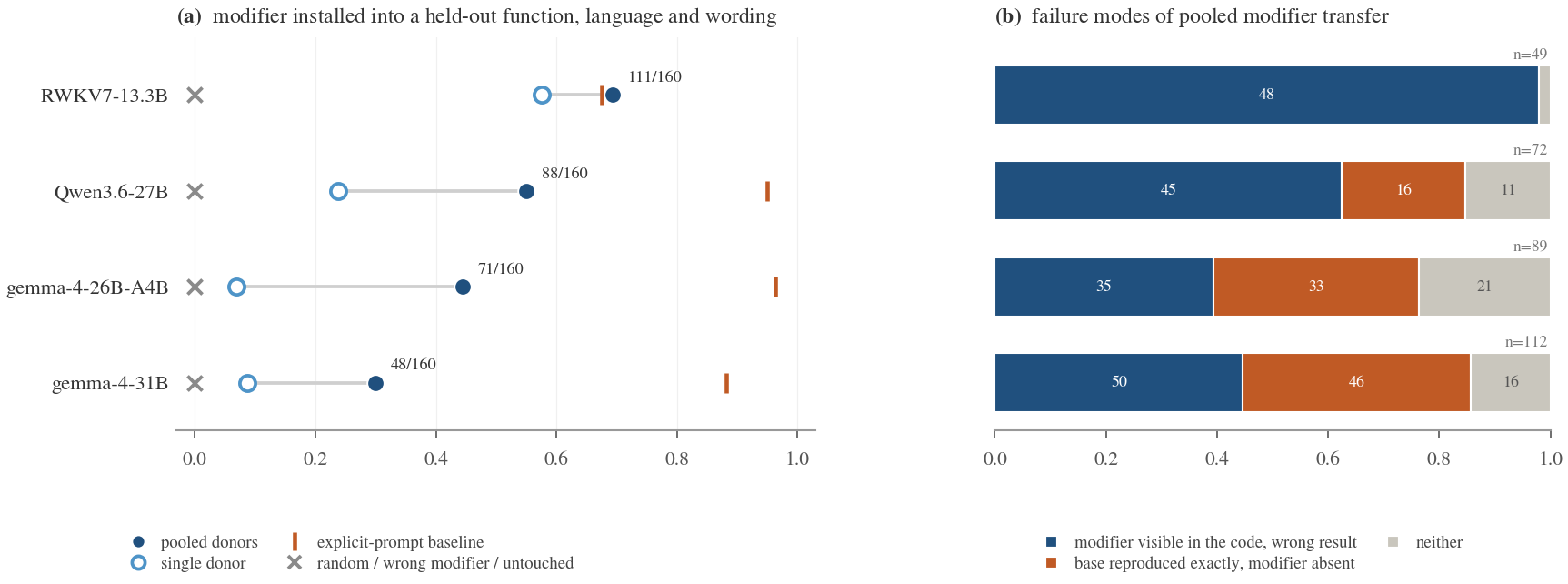}
\caption{
\textbf{Computational modifiers generalize across held-out contexts.}
\textbf{(a)} Target-correct transfer across 160 cells per model.
\textbf{(b)} Outcomes among unsuccessful pooled-donor interventions. Some failures still visibly express the modifier in its erroneous computation. Especially RWKV, which does so in 48 of its 49 misses.}
\label{fig}
\end{figure}

Figure~\ref{fig}a shows modifier transfer in all four models. Pooled estimates succeed in 111/160 cells for \texttt{RWKV7-13.3B}, 88/160 for \texttt{Qwen3.6-27B}, 71/160 for \texttt{Gemma-4-26B-A4B}, and 48/160 for \texttt{Gemma-4-31B}. Pooling improves over a single donor in every model, while none of the random, wrong-modifier, or unmodified controls produce the target computation.

Perhaps most interesting, an unsuccessful intervention does not always mean the intervention was inert. This is clearest for RWKV: of its 49 misses, 48 visibly express the modifier and none simply reproduce the unmodified base (Fig.~\ref{fig}b). Thus 159 of its 160 generations either implement the target or visibly express the intended transformation. The modifier arrives, the model just expresses it incorrectly.

There is a large difference in terms of how prompt information is available during generation. In transformers, the models can continue to use representations at earlier token positions, while RWKV must compress all of its information forward through its recurrent state.

Overall, the result is that computational modifiers \textit{can} transfer across held-out functions, languages, wordings, and models, while remaining substantially more local than a globally linear function representation. However, the transfer also paints a complicated picture and reveals that we do not fully understand the geometry. An important caveat seems to be \textit{where} in the program a certain modifier is expressed and \textit{how} the model keeps track of this.

\subsection{Steering Dosage}
\label{app:dosage}
Here we additionally investigate how sensitive the interventions are to the choice of $\alpha$; the scalar applied to our steering vectors.

\begin{figure}[H]
    \centering
    \includegraphics[width=0.9\linewidth]{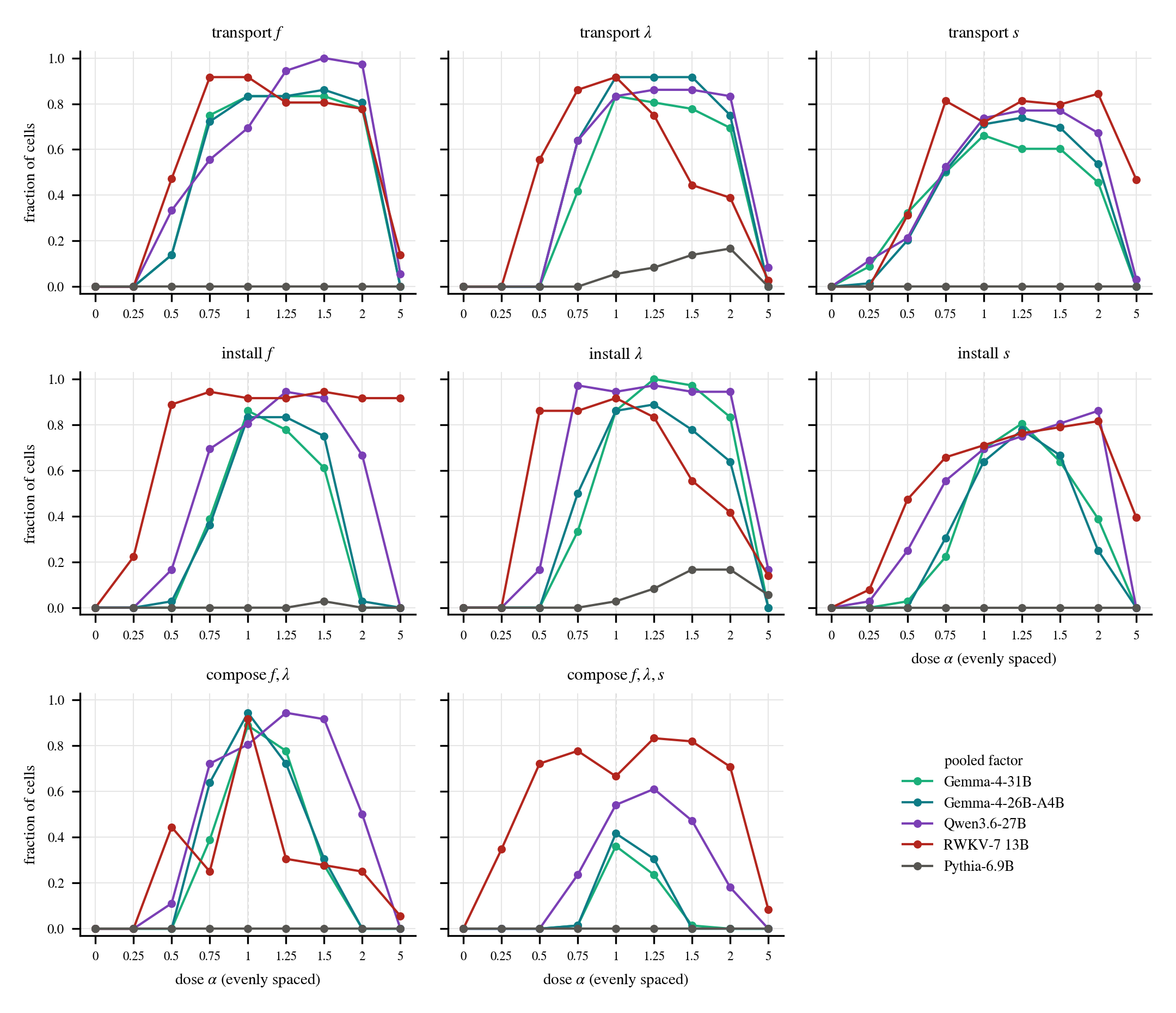}
    \caption{The effectiveness of different steering strengths $\alpha$ for language and function transport and install.}
    \label{fig:dosage}
\end{figure}

Figure \ref{fig:dosage} shows that the intervention is fairly robust to choice of $\alpha$, but that---as expected---performance drops off at more extreme values. Notably, values that are not exactly equal to $1$ are still successful.

\newpage

\subsection{Paraphrases}
\label{app:paraphrase}
\begin{figure}[H]
    \centering
    \includegraphics[width=0.9\linewidth]{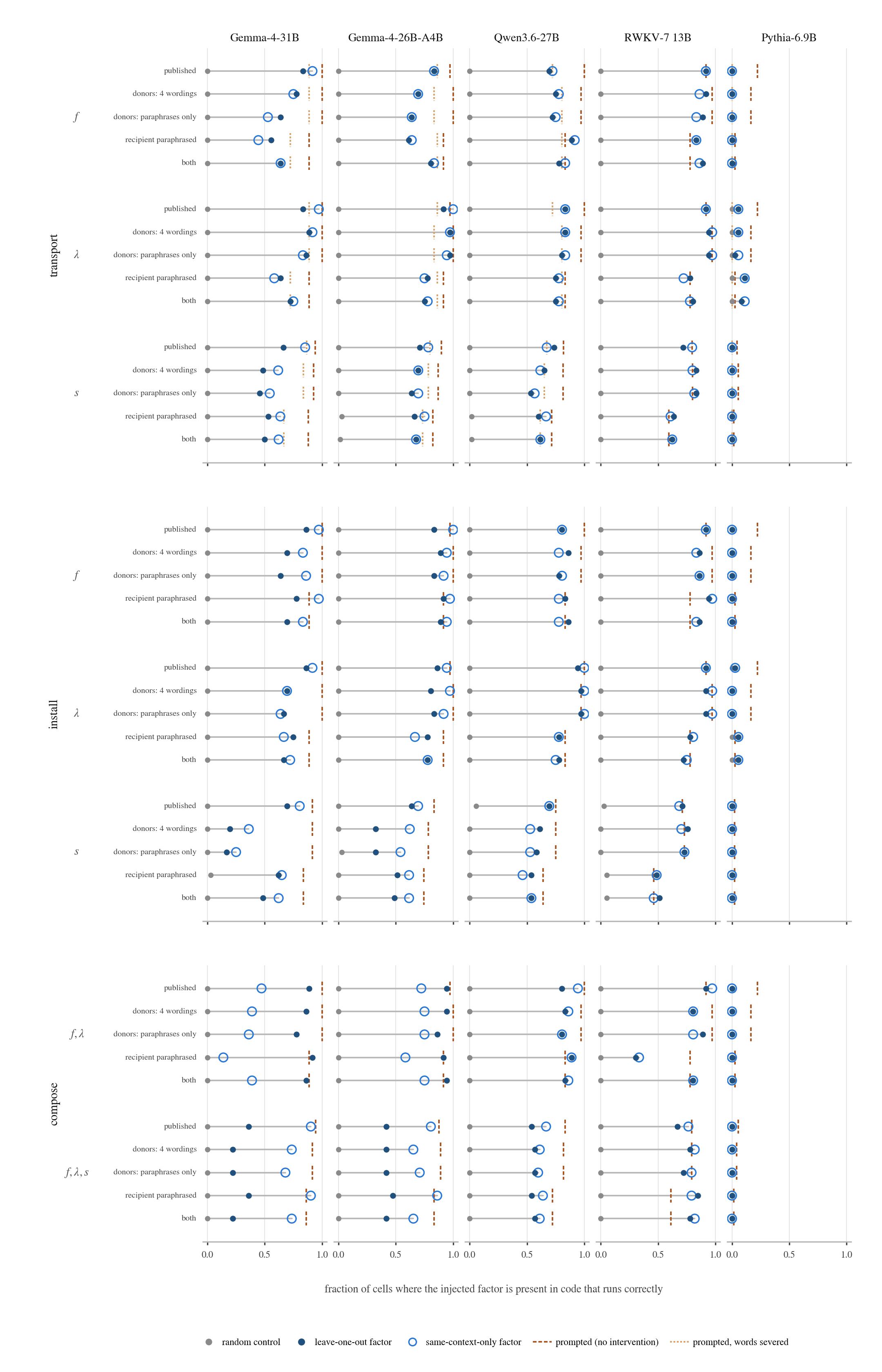}
    \caption{The paraphrasing interventions largely still succeed, albeit at a lower rate than when the descriptions are kept fixed between donor and recipient.}
    \label{fig:paraphrase}
\end{figure}

One objection to our experimental setup is that things are kept so similar between prompts that part of the effect can be attributed to being close to a context-swap. To address this, we investigate 4 additional conditions:

\begin{itemize}
    \item Each donor is averaged over 4 different phrasings, the recipient is one of those.
    \item Donors randomly choose between 3 different phrasings, the recipient is a held-out phrasing.
    \item Donors have fixed phrasing, the recipient randomly chooses between 3 held-out phrasings.
    \item Each donor is averaged over 3 different phrasings, the recipient is a held-out phrasing.
\end{itemize}

Figure \ref{fig:paraphrase} shows that our interventions are largely robust to paraphrasing, but there is a notable decrease in performance, especially under the same-context interventions. These results argue that the interventions succeed not merely by transporting prompt-specific information between nearly identical contexts. The decrease under same-context interventions, however, suggests that the representation is not perfectly invariant to wording, where averaging appears to  reduce this effect. A possible explanation is that averaging suppresses prompt-specific phrasings and isolates a more reusable representation. Future work should seek to understand this effect and determine if there is perhaps some shared causal subspace that averaging helps isolate.

\end{document}